\pdfoutput=1

\documentclass[11pt]{article}

\usepackage[final]{acl}

\usepackage{times}
\usepackage{latexsym}

\usepackage[T1]{fontenc}

\usepackage[utf8]{inputenc}

\usepackage{microtype}

\usepackage{inconsolata}

\usepackage{graphicx}

\usepackage{subcaption}

\usepackage{pifont}

\newcommand{\cmark}{\ding{51}}  
\newcommand{\xmark}{\ding{55}}  

\usepackage{float}

\usepackage{array}

\usepackage{multirow}

\usepackage{amsmath}
\usepackage{amsfonts}

\usepackage{listings}
\usepackage[ruled,vlined]{algorithm2e}

\usepackage{url}
\usepackage{xspace}
\usepackage{caption}
\usepackage{wrapfig}
\usepackage{algpseudocode}
\usepackage{amsmath,amssymb}
\usepackage{float}
\usepackage{graphicx}
\usepackage{csquotes}
\usepackage{xcolor}
\usepackage{multirow}
\usepackage{multicol}
\usepackage{wrapfig,lipsum,booktabs}
\usepackage{makecell}
\usepackage{wrapfig}
\usepackage{array}
\usepackage{color, colortbl}
\usepackage{pifont}
\usepackage{enumitem}
\usepackage{inconsolata}
\usepackage{subcaption}
\usepackage{xcolor}
\usepackage[normalem]{ulem} 
\usepackage{soul}
\newcommand{\ctext}[3][RGB]{%
  \begingroup
  \definecolor{hlcolor}{#1}{#2}\sethlcolor{hlcolor}%
  \hl{#3}%
  \endgroup
}

\usepackage{bm}

\title{Dynamic Rewarding with Prompt Optimization Enables \\ Tuning-free Self-Alignment of Language Models}

\author{%
Somanshu Singla\textsuperscript{$*\clubsuit$} \ 
Zhen Wang\thanks{Equal contribution}\textsuperscript{$\clubsuit$ $\spadesuit$} \
Tianyang Liu\textsuperscript{$\clubsuit$}\  
 \\ 
\textbf{ Abdullah Ashfaq\textsuperscript{$\clubsuit$} \
Zhiting Hu\textsuperscript{$\clubsuit$} \ 
Eric P. Xing\textsuperscript{$\spadesuit$ $\diamondsuit$}}  
\vspace{5pt} \\
\textsuperscript{$\clubsuit$}UC San Diego \
\textsuperscript{$\spadesuit$}MBZUAI \ \textsuperscript{$\diamondsuit$} CMU \\
\texttt{\{ssingla, zhw085\}@ucsd.edu}  
}

\def\ours{\text{DRPO}\xspace}

\definecolor{forest green}{RGB}{34, 139, 34}

\renewcommand{\cmark}{\textcolor{forest green}{\ding{51}}}
\renewcommand{\xmark}{\textcolor{red}{\ding{55}}}

\begin{document}
\maketitle
\begin{abstract}
  
  Aligning Large Language Models (LLMs) traditionally relies on costly training and human preference annotations. Self-alignment seeks to reduce these expenses by enabling models to align themselves. To further lower costs and achieve alignment without any expensive tuning or annotations, we introduce a new tuning-free approach for self-alignment, Dynamic Rewarding with Prompt Optimization (\ours). Our approach leverages a search-based optimization framework that allows LLMs to iteratively self-improve and craft the optimal alignment instructions, all without additional training or human intervention. The core of \ours is a dynamic rewarding mechanism, which identifies and rectifies model-specific alignment weaknesses, allowing LLMs to adapt efficiently to diverse alignment challenges. Empirical evaluations on eight recent LLMs, both open- and closed-sourced, demonstrate that \ours significantly enhances alignment performance, with base models outperforming their SFT/RLHF-tuned counterparts. Moreover, the prompts automatically optimized by \ours surpass those curated by human experts, further validating the effectiveness of our approach. Our findings highlight the great potential of current LLMs to achieve adaptive self-alignment through inference-time optimization, complementing tuning-based alignment methods.\footnote{Code available: \url{https://github.com/Singla17/DRPO}}
  
\end{abstract}

\section{Introduction}

Aligning Large Language Models (LLMs, ~\citealt{brown2020language,chowdhery2023palm, touvron2023llama,achiam2023gpt}) with human ethical standards and practical expectations is extremely crucial to prevent unintended consequences and ensure AI's positive contribution to society. Traditional alignment methods, such as supervised fine-tuning (SFT) and reinforcement learning from human feedback (RLHF)~\cite{bai2022constitutional, ouyang2022training}, are resource-intensive and require extensive human oversight, limiting their scalability and practicality. As LLMs grow more complex and widely adopted, the demand for cost-effective, annotation-efficient, and rapidly adaptable alignment strategies becomes increasingly urgent.

\begin{figure}
    \centering
    \includegraphics[width=1\linewidth]{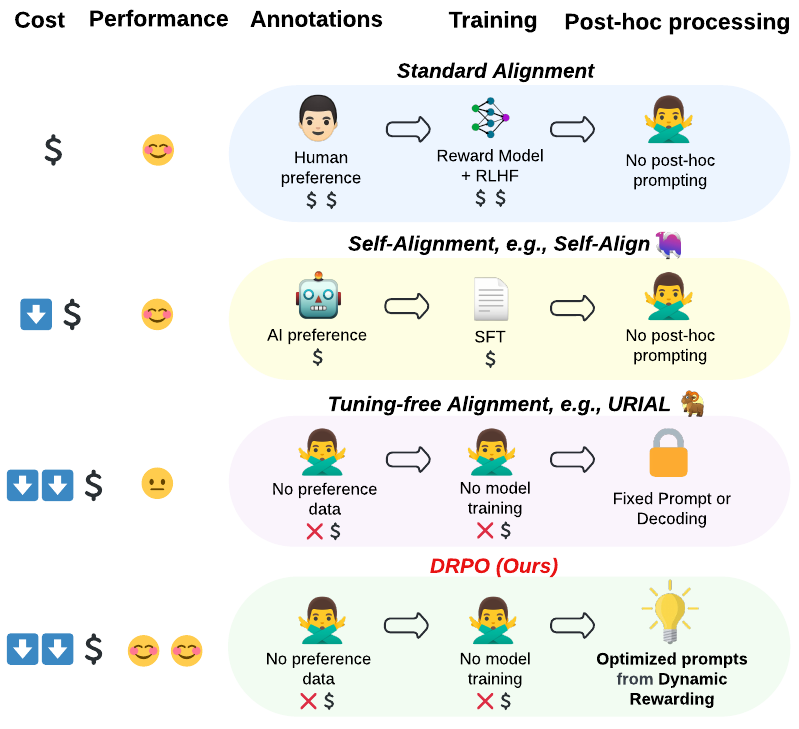}
    \vspace{-18pt}
    \caption{Comparison of \ours with other LLM alignment paradigms. \ours combines the benefits of self-alignment and tuning-free alignment, enabling self-improvement and high cost-efficiency without requiring human supervision or additional model training.
    }
    \vspace{-22pt}
    \label{fig:paradigm_comparison}
\end{figure}

Self-alignment aims to improve LLM alignment by leveraging the models themselves; for example, by replacing human feedback with model-generated feedback~\cite{lee2023rlaif}, synthesizing preference data~\cite{kim2023aligning, sun2024principle}, or self-critique~\cite{bai2022constitutional}. Despite these advancements, such methods still demand significant resources, including the costly and unstable RLHF tuning, as well as some level of human supervision, such as carefully curated alignment rules or in-context learning (ICL) prompts~\cite{sun2024principle}. On the other hand, as shown in Figure~\ref{fig:paradigm_comparison}, a recent line of research focuses on tuning-free alignment, which prioritizes extreme efficiency without incurring any tuning cost. These approaches include techniques like decoding-based alignment~\cite{li2023rain, wang2024inferaligner} or ICL alignment~\cite{han2023context, Lin2024ReAlign, zhao2024context}. However, these tuning-free methods are often static (e.g., relying on fixed prompts or reward functions) and thus lack the flexibility to adapt and self-improve for better alignment.

To marry the strengths of both paradigms, in this paper, we propose \ours, Dynamic Rewarding with Prompt Optimization, a novel tuning-free approach for LLM self-alignment. \ours draws inspiration from two key insights from recent alignment research. First, the superficial alignment hypothesis~\cite{zhou2024lima} suggests that LLMs can be effectively aligned through lightweight tuning or even simple prompting~\cite{Lin2024ReAlign, zhao2024context}. Second, reward models in RLHF often generalize poorly to out-of-distribution samples~\cite{burns2023weak}, whereas LLMs, known for their superior generalization capabilities, can provide more effective rewards and feedback for alignment purposes. Building on these insights, \ours is constructed atop a search-based prompt optimization (PO) framework~\cite{pryzant2023automatic, hao2023reasoning, wang2023promptagent}, which enables LLMs to self-correct and automatically craft detailed alignment instructions. This steers model behavior more effectively, without relying on any use of human preferences or model training.

The core novelty of \ours lies in its \textit{dynamic rewarding} mechanism, integrated with the optimization framework. This mechanism allows LLM-based rewards to be dynamically adjusted based on specific queries, helping to identify and address the model's alignment blind spots. For example, if an LLM with outdated knowledge pretends to answer a question requiring the latest news, its ``knowledge limitation'' reward will be low, and the alignment prompt will be updated accordingly. We apply this novel method to automatically craft both the system prompt and responses in ICL examples, which have proven highly effective in improving alignment.

We conducted comprehensive experiments on 8 recent LLMs using the standard alignment benchmark, \texttt{just-eval-instruct}, composed of questions from multiple alignment datasets. Our results show that \ours can effectively align both base and SFT/RLHF tuned models. Notably, \ours significantly enhances base models, enabling them to outperform their SFT/RLHF-tuned counterparts. \ours can further improve SFT/RLHF-tuned models, highlighting its compatibility with other tuning-based alignment techniques. Additionally, our automatically optimized prompts substantially outperform those curated by human experts.

\begin{figure}
    \centering
    \includegraphics[width=1.0\linewidth]{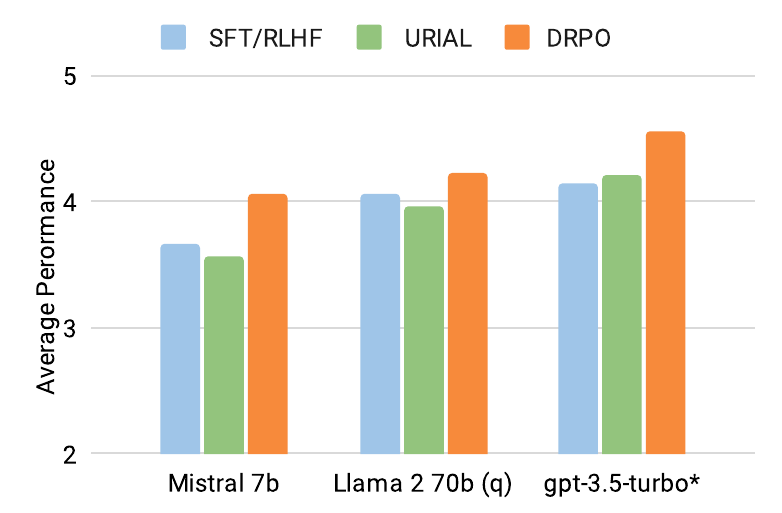}
    \vspace{-15pt}
    \caption{Comparison of \ours with other alignment methods, including RLHF and URIAL~\cite{Lin2024ReAlign}. \ours consistently outperforms both baselines across multiple LLMs.
    Note that we do not have access to \texttt{gpt-3.5-turbo} base model; hence, both \ours and URIAL are directly applied to its RLHF-tuned version.}
    \label{fig:overall_comparison_chart}
    \vspace{-15pt}
\end{figure}

\section{Related Work}
\vspace{-5pt}

\noindent \textbf{Self-Alignment.}
Traditional alignment approaches rely heavily on extensive human-annotated preference data and complex reward model training through reinforcement learning, which poses significant scalability and cost challenges~\cite{ouyang2022training}. Self-alignment focuses on aligning LLMs themselves with model-generated feedback, datasets, critique, etc., which are then used for fine-tuning or training reward models~\cite{lee2023rlaif, bai2022training, cao2024towards, wang2024step, guo2024human}. Notable examples include synthesizing alignment training data with human-provided instructions and ICL examples~\cite{wang2022self, kim2023aligning, sun2024principle}, augmented web documents~\cite{li2023self}, or self-critique~\cite{bai2022constitutional, madaan2024self}. However, most of these methods still require an SFT/RLHF-tuning process to enhance alignment, along with some degree of human annotations or supervision. In contrast, \ours shares similar self-alignment principles using self-critique error feedback to gradually align the model, but it achieves this entirely without any model tuning or human supervision.

\begin{figure*}[ht]
    \centering
    \includegraphics[width=0.95\linewidth]{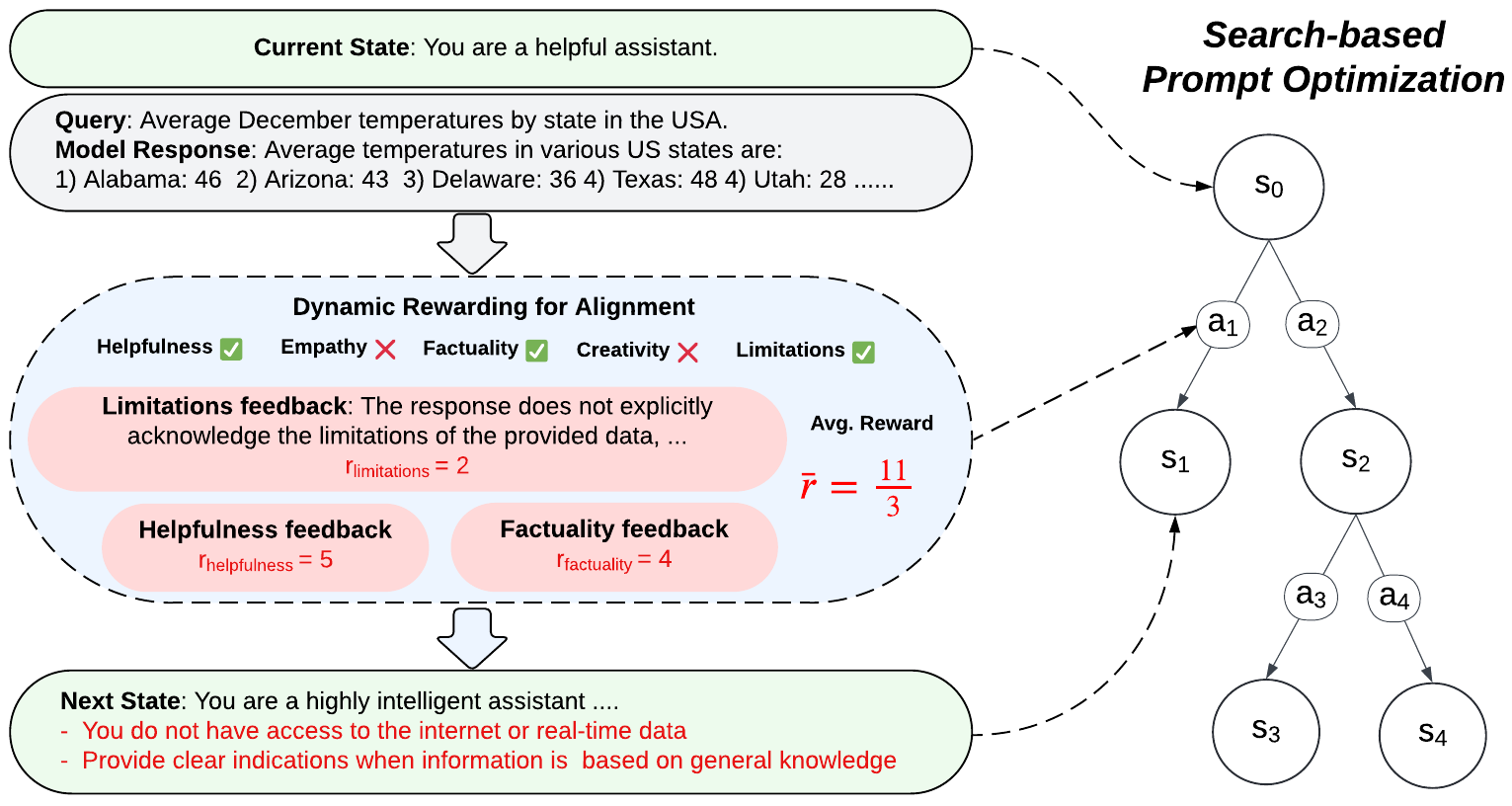}
    \vspace{-8pt}
    \caption{Overall framework of Dynamic Rewarding with Prompt Optimization (\ours). The optimization problem is modeled as a Markov Decision Process (MDP) and solved using beam search to optimize the alignment prompt. Dynamic rewarding, a novel technique integrated into this framework, allows flexible reward assignment to detect and address alignment weaknesses in the current LLM, thereby enhancing the overall optimization process.
    }
    \label{fig:dynamic_rewarding}
    \vspace{-15pt}
\end{figure*}

\noindent \textbf{Tuning-Free Alignment.}
A recent trend in alignment research is to align LLMs without updating their parameters, typically as a post-training process for LLMs. This has witnessed two major lines of work recently. The first aligns models with carefully curated human annotations and ICL examples~\cite{han2023context, Lin2024ReAlign, zhao2024context}, while the second involves decoding-based methods to guide token generation and search with alignment rewards~\cite{li2023rain, khanov2024args, huang2024deal}. Although tuning-free, the first approach still requires human curation and often underperforms compared to SFT/RLHF-tuned counterparts. The second one, while effective, incurs higher inference costs per query, making it computationally expensive. It is worth mentioning that another recent promising direction is cost-efficient alignment through representation engineering~\cite{zou2023representation, wu2024reft}, which aims to steer LLM representation vectors for alignment~\cite{li2024inference, kong2024aligning, wang2024inferaligner}. However, these methods are not fully tuning-free and typically require additional data or model training to identify alignment directions in the embedding space. Nevertheless, \ours requires no additional annotations or model training, and also only needs a one-time optimization per model to achieve better performance than SFT/RLHF-tuned counterparts.

\noindent \textbf{Prompt Optimization.}
Discovering optimal discrete prompts becomes far more crucial nowadays. Modern prompts for LLMs can be generally divided into two parts: in-context learning examples and detailed instructions. The former is usually treated as a retrieval problem with various schemas to select the influential examples~\cite{rubin2021learning, dong2022survey}. Optimizing the latter has been heavily studied recently, mostly formulated as a sampling or search problem. Generally, an initial prompt (e.g., a base prompt, ``You are a helpful assistant'') is given to start an iterative process, where diverse prompt candidates are generated per turn, and the best ones are kept for the next iteration. Various sampling strategies are proposed to diversify the prompt candidates, e.g., back translation~\cite{xu2022gps}, evolutionary operations~\cite{fernando2023promptbreeder}, self-critique~\cite{wang2023promptagent}. Different search frameworks also have been studied, such as Monte Carlo search~\cite{zhou2022large}, evolutionary algorithms~\cite{fernando2023promptbreeder, yang2023large}, beam search~\cite{pryzant2023automatic}, and Monte Carlo tree search (MCTS)~\cite{wang2023promptagent}. \ours builds upon recent search-based optimization methods but introduces novel techniques, such as dynamic rewarding, to effectively address the alignment problem. 
\section{Methodology}
\vspace{-5pt}

In this section, we introduce our formulation formally and present \ours for solving the alignment problem by optimizing the alignment instruction.



\subsection{Problem Formulation}

Given an LLM $\mathcal{B}$, an alignment instruction consists of two parts: a system prompt $\mathcal{P}$ and a set of $N$ in-context learning (ICL) examples $\mathcal{I}$.
The system prompt $\mathcal{P}$ serves as a prefix that provides high-level instructions, sets the tone, and imposes constraints on the model's responses. Each ICL example $\mathcal{I}_i$ consists of a pair $(q_i, d_i)$, where $q_i$ is an input query and $d_i$ is the corresponding desired response, so we can represent $\mathcal{I} = \{(q_1, d_1), (q_2, d_2), \ldots, (q_N, d_N)\}$. 

Conditioning on the system prompt $\mathcal{P}$ and a selected subset of $K$ ICL examples $\mathcal{I}_K \subseteq \mathcal{I}$, the aligned model response $y$ to an input $x$ is generated as:
\[
y = \mathcal{B}(x \mid \mathcal{P}, \mathcal{I}_K)
\]

\ours aims to optimize both system prompt $\mathcal{P}$ and ICL examples $\mathcal{I}_K$ to enhance alignment. This involves finding the best possible $\mathcal{P}^*$ and $\mathcal{I}_K^*$ that maximize the alignment of the model's responses. This optimization problem can be formulated as follows:
\[
(\mathcal{P}^*, \mathcal{I}_K^*) = \arg\max_{\mathcal{P}, \mathcal{I}_K} \mathbb{E}_{x \sim \mathcal{D}_x} \left[\mathcal{B}(x \mid \mathcal{P}, \mathcal{I}_K) \right]
\]
\noindent where $\mathcal{D}_x$ denotes the distribution of input queries, and the expectation $\mathbb{E}$ represents the alignment performance for responses based on specific metrics.




\subsection{Dynamic Rewarding with Prompt Optimization (\ours)}

Given the distinct nature of the system prompt and ICL examples, we propose to optimize them separately, resulting in a two-step optimization approach. We first construct a universal set of ICL examples and optimize their responses to obtain $\mathcal{I}^*$. Next, we estimate a model-specific system prompt $\mathcal{P}^*$ based on the optimized universal set $\mathcal{I}^*$. Notably, we leverage the \texttt{LLM Reasoners}\footnote{\url{https://github.com/maitrix-org/llm-reasoners}} framework~\cite{hao2023reasoning, hao2024llm} as the prompt optimization (PO) framework. Specifically, \texttt{LLM Reasoners} incorporates a base model $\mathcal{B}$, an optimizer $\mathcal{O}$, and an evaluator $\mathcal{E}$. It operates as a search agent that iteratively interacts with the model's environment, using the optimizer $\mathcal{O}$ to adjust the prompt $\mathcal{P}$ or ICL examples $\mathcal{I}$ based on a reward function $\mathcal{R}$. For further details, we refer readers to the original references. In the following, we introduce the core component of \ours.

\subsubsection{Dynamic Rewarding for Alignment}

We formulate this optimization problem as a Markov Decision Process (MDP). In this framework, the states $s\in \mathcal{S}$ represent our optimization goal, which could be either a system prompt or an in-context example. Actions $a \in \mathcal{A}$ are defined based on the alignment feedback obtained during the evaluation of any given state. The key motivation is to leverage the superior generalization capabilities of LLMs to evaluate and analyze states, guiding state transitions toward an optimal state. We employ different evaluation techniques for system prompt and in-context example optimization, which are detailed in subsequent sections. Efficient traversal of this state space is crucial, and for this purpose, we adopt beam search due to its effectiveness and low computational cost.

One of the key challenges in our optimization task is designing a reward function capable of handling a problem as broad and generalized as alignment. As illustrated in Figure~\ref{fig:dynamic_rewarding}, a single, unified reward function is impractical due to the vast query space we aim to align with the base LLM $\mathcal{B}$. Different queries emphasize different focal points, meaning that certain evaluation criteria might be appropriate for some queries but not for others. To overcome this, we introduce a dynamic reward function $\mathcal{R}$, which can dynamically adapt to the specific query being evaluated. Notably, our approach shares conceptual similarities with a few recent alignment research, which also advocate for adaptable and query-sensitive alignment strategies~\cite{bai2022constitutional, sun2024principle}. However, the key distinction lies in our dynamic reward function’s ability to not only enable flexible evaluation but also integrate seamlessly into a formally defined optimization framework.

Specifically, we first predefined a set of reward criteria $\mathbb{R}$, from which the model dynamically selects the most relevant rewards, while also retaining the flexibility to propose new ones when necessary. Formally, for a given query \( q \), the dynamic reward function $\mathcal{R}$ evaluates the model's response $\sigma$ based on a dynamically selected or proposed rewards $\mathbb{R}_q$, where $\mathbb{R}_q \subseteq \mathbb{R} \cup \mathbb{R}^*$ and $\mathbb{R}^*$ represents newly proposed rewards. The reward function is defined as:

\[
\mathcal{R}(\sigma \mid \mathbb{R}_q) = \frac{1}{|\mathbb{R}_q|} \sum_{r \in \mathbb{R}_q} r(\sigma)
\]

Here, $\mathbb{R}_q$ denotes relevant rewards tailored for the given query \( q \) and \(r(\sigma)\) denotes the score of a specific reward when evaluating any response \(\sigma\). 

This allows us to flexibly score and evaluate responses based on the most relevant criteria for each specific query, ensuring that the evaluation remains contextually appropriate and comprehensive.

\subsubsection{ICL Example Optimization}

To optimize in-context learning examples, we start with a set of base ICL examples $\mathcal{I}_{\text{base}} = \{(q_1, b_1), (q_2, b_2), \ldots, (q_N, b_N)\} $, where $q_i$ is a query and $b_i$ is a base response to the query, $N$ is the total number of in-context examples. Our overall goal is to find a universal set $\mathcal{I}^{*}$ that maximizes alignment across various models.

We specifically optimize each ICL example $(q_i, b_i)$ individually. The initial state of the search tree for an ICL example is defined as the base response to the query, i.e.,  $s_0 = b_i$. At any time $t$, the state of the search tree, $s_t$, is the response of the example. This allows us to systematically monitor and evaluate the response at any given time $t$. The state space $\mathcal{S}$ encompasses all possible responses to the query $q_i$.

To evaluate and improve the alignment, we use the dynamic reward function $\mathcal{R}$. The relevant rewards $\mathbb{R}_{q_i}$ for the query $q_i$ are specifically selected or potentially proposed new rewards. The reward function $\mathcal{R}$ and evaluator $\mathcal{E}$ then evaluates the state $s_t$ based on these rewards, providing a reward $r_t$ and alignment feedback $a_t$:

\[
\begin{aligned}
& r_t = \mathcal{R}(s_t \mid \mathbb{R}_{q_i}) \\
& a_t = \mathcal{E}(s_t \mid \mathbb{R}_{q_i})
\end{aligned}
\]

Note that, in practice, evaluation and reward generation are performed simultaneously using one single prompt, so the evaluation can also be considered dynamic. The transition function $\mathcal{T}$, implemented by optimizer $\mathcal{O}$, then updates the state:
\[
s_{t+1} = \mathcal{T}(s_t, a_t)
\]

The detailed pseudo-code for this optimization process is provided in Algorithm \ref{alg:icl_opti} in Appendix \ref{sec:opti_algo} and the prompts used by our algorithm can be found in Appendix \ref{sec:meta_prompts}.

\subsubsection{System Prompt Optimization}

The optimization process for the system prompt is similar to that of the ICL example optimization. For the system prompt optimization, we use $K$ optimized ICL examples $\mathcal{I}_K^*  \subseteq \mathcal{I}^*$, where the $K$ ICL examples are chosen using similarity-based retrieval. We collect a set of seed samples $\mathcal{X} = \{x_1, x_2, \ldots, x_N \}$, where $x_i$ is a query that will be used to test the alignment of the base model $\mathcal{B}$. The goal of this process is to find the optimal prompt $\mathcal{P}^*$ (given that we already have access to $\mathcal{I}_K^*$), such that alignment of LLM $\mathcal{B}$ is maximized. This prompt is specific to the base model $\mathcal{B}$ and will provide the model with actionable insights and guidance to improve its alignment.

The optimization process begins by defining the initial state $s_0$ as the basic system prompt (e.g., ``You are a helpful assistant.''). At any time $t$, the state $s_t$ represents the current system prompt, and the state space $\mathcal{S}$ includes all possible system prompts for the given LLM $\mathcal{B}$.

For a given state $s_t$, we sample a query $x_t$ from the seed samples $\mathcal{X}$. The relevant rewards $\mathbb{R}_{x_t}$ for the query $x_t$ are specifically selected or potentially proposed new rewards. The reward function $\mathcal{R}$ and the evaluator $\mathcal{E}$ then evaluate the response generated by the model $\mathcal{B}$ given the system prompt $s_t$ and the selected in-context examples $\mathcal{I}_K^*$, providing a reward $r_t$ and alignment feedback $a_t$:

\[
\begin{aligned}
& r_t = \mathcal{R}(\mathcal{B}(x_t \mid s_t, \mathcal{I}_K^*)\mid \mathbb{R}_{x_t}) \\
& a_t = \mathcal{E}(\mathcal{B}(x_t \mid s_t, \mathcal{I}_K^*)\mid \mathbb{R}_{x_t})
\end{aligned}
\]

The optimizer $\mathcal{O}$ as a transition function then updates the state, $ s_{t+1} = \mathcal{T}(s_t, a_t) $. The detailed pseudo-code for this optimization process is provided in Algorithm \ref{alg:prompt_opti} in Appendix \ref{sec:opti_algo}.
\section{Experiments}

\subsection{Experimental Setup}


\begin{table*}[t]
\begin{center}
\begin{tabular}{ >{\raggedright\arraybackslash}p{3.9cm} c c c c c c c c } 
    \toprule

    \textbf{[Tuned] Model} & \textbf{Method }& \bm{$K$} & \textbf{Helpful} & \textbf{Clear} & \textbf{Factual} & \textbf{Deep} & \textbf{Engage} & \textbf{Avg.} \\
    \midrule

    [\xmark] Mistral 7b  & Base & 0 & 2.20 & 2.51 & 2.29 & 1.69 & 1.80 & 2.10 \\
    
   [\xmark] Mistral 7b  & URIAL & 3 & 3.62 & 4.32 & 3.75 & 2.70 & 3.41 &  3.56\\
    
    [\xmark] Mistral 7b  & \ours & 2 & \textbf{4.23} & \textbf{4.56} & \textbf{3.97} & \textbf{3.68} & \textbf{3.84} &  \textbf{4.06}\\
    
    \hline
    
   [\cmark] Mistral 7b (Instruct) & Base & 0 & 3.98 & 4.44 & 3.64 & 2.97 & 3.26 &  3.66\\
    
   [\cmark] Mistral 7b (Instruct) & URIAL & 3 & 3.94 & 4.51 & 3.69 & 2.99 & 3.75 &  3.78\\
    
    [\cmark] Mistral 7b (Instruct) & \ours & 2 & \textbf{4.22} & \textbf{4.60} & \textbf{3.80} & \textbf{3.68} & \textbf{3.99} &  \textbf{4.06}\\

   \hline

    [\xmark] Llama 2 70b$^q$  & Base & 0 & 2.07 & 2.55 & 2.35 & 1.50 & 1.63 &  2.02 \\
    
    [\xmark] Llama 2 70b$^q$ & URIAL & 3 & 4.25 & 4.67 & 4.03 & 3.08 & 3.80 &  3.97 \\
    
    [\xmark] Llama 2 70b$^q$  & \ours & 2 & \textbf{4.42} & \textbf{4.72} & \textbf{4.23} & \textbf{3.81} & \textbf{3.98} &  \textbf{4.23}\\

    \hline

    [\cmark] Llama 2 70b$^q$ (chat) & Base & 0 & 4.36 & 4.71 & 3.95 & 3.56 & 3.76 &  4.07\\

    [\cmark] Llama 2 70b$^q$ (chat) & URIAL & 3 & 4.32 & 4.72 & 4.08 & 3.50 & 4.25 &  4.17\\
    
    [\cmark] Llama 2 70b$^q$ (chat) & \ours & 2 & \textbf{4.46} & \textbf{4.75} & \textbf{4.10} & \textbf{4.11} & \textbf{4.37} &  \textbf{4.36}\\

   \hline

    [\xmark] Llama 3 8b  & Base & 0 & 1.82 & 2.27 & 2.20 & 1.38 & 1.48 &  1.83\\

    [\xmark] Llama 3 8b  & URIAL & 3 & 3.94 & \textbf{4.51} & 3.69 & 2.99 & \textbf{3.75} & 3.78 \\
    
    [\xmark] Llama 3 8b  & \ours & 2 & \textbf{4.02} & 4.40 & \textbf{3.84} & \textbf{3.50} & 3.65 &  \textbf{3.88} \\

   \hline

    [\cmark] Llama 3 8b (Instruct) & Base & 0 & 4.43 & 4.72 & 3.98 & 3.45 & 3.76 &  4.07\\

    [\cmark] Llama 3 8b (Instruct) & URIAL & 3 & 4.48 & 4.81 & \textbf{4.19} & 3.55 & 4.27 &  4.26\\
    
    [\cmark] Llama 3 8b (Instruct) & \ours & 2 & \textbf{4.54} & \textbf{4.81} & 4.16 & \textbf{4.08} & \textbf{4.40} & \textbf{4.40} \\

    \hline

    [\cmark] \texttt{gpt-3.5-turbo} & Base & 0 & 4.56 & 4.89 & 4.41 & 3.30 & 3.55 & 4.14 \\

    [\cmark] \texttt{gpt-3.5-turbo} & URIAL & 3 & 4.30 & 4.77 & 4.41 & 3.44 & 4.11 &  4.21\\
    
    [\cmark] \texttt{gpt-3.5-turbo} & \ours & 2 & \textbf{4.67} & \textbf{4.92} & \textbf{4.53} & \textbf{4.07} & \textbf{4.58} &  \textbf{4.55}\\

   \hline
    [\cmark] \texttt{gpt-4-0613} & Base & 0 & \textbf{4.71} & \textbf{4.93} & \textbf{4.52} & 3.49 & 3.53 &  \textbf{4.24} \\

    \bottomrule

\end{tabular}

\caption{Performance on \texttt{just-eval-instruct} benchmark. ``Tuned'' indicates whether the model has been SFT/RLHF tuned. Models are evaluated across multiple aspects: ``Helpful'' (Helpfulness), ``Clear'' (Clarity), ``Factual'' (Factuality), ``Deep'' (Depth), and ``Engage'' (Engagement). The base method indicates a basic alignment prompt. Our method consistently outperforms baseline methods across multiple aspects and overall.}
\label{tab:main_table}
\vspace{-17pt}
\end{center}
\end{table*}

\noindent \textbf{Evaluation Dataset}.
We use the standard alignment benchmark, \texttt{just-eval-instruct}~\cite{Lin2024ReAlign}, which merges five popular alignment datasets to provide a comprehensive and fine-grained evaluation of LLM alignment. This benchmark consists of 1,000 examples: the first 800 assess the models' helpfulness, and the remaining 200 evaluate their harmlessness. The first 800 examples are evaluated based on five fine-grained aspects: \textit{helpfulness}, \textit{clarity}, \textit{factuality}, \textit{depth}, and \textit{engagement}, while the remaining 200 are evaluated using the \textit{safety} aspect. We use GPT-4 Turbo (\texttt{gpt-4-1106-preview}), one of the latest GPT-4 models available during our experiments, to evaluate both types of examples using the prompts specified in the original URIAL paper~\cite{Lin2024ReAlign}. The scoring scale ranges from 1 to 5, indicating ``strongly disagree'', ``disagree'', ``neutral'', ``agree'', and ``strongly agree''. Note that we employ a more recent version of GPT-4 compared to URIAL, which enhances the strictness and accuracy of our evaluation pipeline. Thus, we re-benchmark URIAL under our updated evaluation setting for consistency across all results.

\noindent \textbf{Seed Samples}. 
When optimizing the system prompt with \ours, we sample from our seed dataset $\mathcal{X}$ to measure the alignment performance of the system prompt at each time step. This seed dataset, consisting of 180 examples, is built using data from \texttt{AlpacaEval} \cite{alpaca_eval}, \texttt{LIMA} \cite{zhou2024lima}, and \texttt{HH-RLHF-redteam} \cite{Ganguli2022RedTL}. More details about the construction of this dataset can be found in Appendix \ref{sec:impl_details}.

\noindent \textbf{Models}.
We benchmark 6 open-source LLMs in our experiments: Mistral 7b (v0.1), Mistral 7b (Instruct)~\cite{Jiang2023Mistral7}, Llama 2 70$b^q$, Llama 2 70$b^q$ (chat) (4-bit AWQ~\cite{lin2023awq} quantized models)~\cite{Touvron2023Llama2O}, Llama 3 8b, Llama 3 8b (Instruct)~\cite{llama3modelcard} and 2 closed-source models: OpenAI's GPT-3.5 Turbo (\texttt{gpt-3.5-turbo}) and GPT-4 (\texttt{gpt-4-0613}). Models without the ``chat'' or ``instruct'' tag are base models, i.e., not tuned by SFT/RLHF. For evaluation, we use greedy decoding (temperature = 0) to ensure reproducibility.

\noindent \textbf{Baselines}. 
We first apply \ours to the base model, making the SFT/RLHF-tuned counterparts without \ours a natural baseline. For instance, we compare Mistral 7B + \ours and Mistral 7b (Instruct). Additionally, we have two more baselines: (1) The base method, where a basic prompt is applied without using ICL examples. (2) URIAL~\cite{Lin2024ReAlign}, where we use the prompt and ICL examples proposed by authors. We also provide extensive ablation baselines of our method, such as changing the search algorithm from Beam search to Greedy Search or Monte Carlo search and using ``static rewarding'' to understand the impact of dynamic rewarding. Full details of these can be found in Appendix~\ref{sec:impl_details}.

\noindent \textbf{Implementation details}.
We use GPT-4-turbo (\texttt{gpt-4-0125-preview}) as both the optimizer $\mathcal{O}$, and evaluator $\mathcal{E}$ unless specified otherwise. The initial set of in-context learning examples, $\mathcal{I}_{base}$, contains 16 examples: 3 from URIAL \cite{Lin2024ReAlign} and 13 generated using \texttt{gpt-4-0125-preview}. More details about the design choice made for $\mathcal{I}_{base}$ can be found in Appendix \ref{sec:impl_details}. We employ sentence transformers \cite{reimers-2019-sentence-bert} to retrieve K in-context learning examples from $\mathcal{I}^*$ given the query. We use $D$ as the beam depth, $W$ as the beam width, and $M$ as the number of action samples per state (to grow the tree for the next iteration). The exact hyper-parameters can be found in Appendix \ref{sec:impl_details}.

\subsection{Results}

\noindent \textbf{Comparison with baselines}. 
Table \ref{tab:main_table} presents the performance comparison of \ours with baselines. \ours outperforms all baselines across both tuned and un-tuned models. As shown in Figure \ref{fig:overall_comparison_chart} using \ours on strong base models such as Mistral 7b and LLama 2 70b$^q$ can surpass even the RLHF/SFT tuned models under base setting. It is noteworthy that \ours achieves superior performance compared to URIAL \citep{Lin2024ReAlign}, despite using fewer in-context learning examples, highlighting the quality of optimized alignment instruction by \ours. Note that while \texttt{just-eval-instruct} includes a \textit{safety} metric, we are not reporting it because, in our analysis, we found that the safety metric is saturated, with all methods (RLHF/SFT, URIAL, and \ours) achieving consistently high scores. This saturation is a good sign, demonstrating that tuning-free methods like \ours can result in very safe models that adhere to human values.

\noindent \textbf{Categorized performance}. 
Appendix~\ref{sec:cat_perf} presents the performance of models across various domains, e.g., ``procedure'', ``lifestyle'', ``info-seek'', ``STEM'', etc. In this experiment, we apply \ours to base models and compare their performance across multiple human-relevant and alignment-critical domains. \ours demonstrates consistently strong performance, surpassing RLHF/SFT-tuned models in most domains across all baselines.

\begin{table}[!t]
\begin{center}
\begin{tabular}{ c c c c } 
    \toprule
    \multirow{2}{*}{\textbf{Model}} & \textbf{Mistral}  & \textbf{Llama}  & \textbf{Base} \\
    & \textbf{Prompt} & \textbf{Prompt} & \textbf{Prompt} \\
    \midrule
    Mistral 7b & \textbf{4.06} & 4.03 & 4.04 \\
    Llama 2 70$b^q$ & 4.19 & \textbf{4.23} & 4.17 \\
    \bottomrule
\end{tabular}
\caption{Effect of prompt transfer on base LLMs. The best performance is achieved when using a prompt specifically optimized for the target base LLM.}
\label{tab:prompt_transfer}
\vspace{-17pt}
\end{center}
\end{table}
\noindent \textbf{Prompt transfer}. 
We also conduct experiments on prompt transfer, i.e., evaluating the performance of an alignment instruction optimized for one LLM on a different LLM. Table~\ref{tab:prompt_transfer} presents the results of transferring various optimized prompts to Mistral 7b and Llama 2 70$b^q$. While the best results are achieved with prompts specifically optimized for the target model, transferring an optimized prompt can still lead to significant alignment improvements. This is evident in the case of LLaMA 2 70B$^q$, which benefits from the prompt optimized for Mistral 7B.

\noindent \textbf{Ablation on system prompt and  ICL examples}. 
Table \ref{tab:ablation_icl_prompt} shows the effect of ablating system prompt and in-context learning examples from \ours. Using both system prompt and in-context learning examples gave the best performance, underscoring the importance of both in alignment. It is worth pointing out that performance degradation on the removal of in-context learning examples was higher when compared to the removal of the system prompt, hinting that in-context learning examples are relatively important in alignment. Given this, our optimized in-context learning examples are a valuable asset and will be released publicly to facilitate further alignment research\footnote{\url{https://github.com/Singla17/DRPO}}.

\begin{table}[!t]
\begin{center}
\resizebox{0.9\linewidth}{!}{%
\begin{tabular}{ c c c c } 
 \toprule
 \multirow{2}{*}{\textbf{Model}} & \textbf{System} & \textbf{ICL} & \multirow{2}{*}{\textbf{Avg.}} \\
 & \textbf{Prompt} & \textbf{(}\bm{$K = 2$}\textbf{)} &  \\
 \midrule

  Mistral 7b  & \cmark & \cmark & \textbf{4.06} \\
 Mistral 7b (Instruct) & \cmark & \cmark & \textbf{4.06} \\
 Llama 2 70$b^q$  & \cmark & \cmark & \textbf{4.23} \\
 \texttt{gpt-3.5-turbo} & \cmark & \cmark & \textbf{4.55} \\

 \midrule

  Mistral 7b & \xmark & \cmark & 4.04 \\
 Mistral 7b (Instruct) & \xmark & \cmark & 4.04 \\
 Llama 2 70$b^q$  & \xmark & \cmark & 4.17 \\
 \texttt{gpt-3.5-turbo} & \xmark & \cmark & 4.42 \\

  \midrule

 Mistral 7b (Instruct) & \cmark & \xmark & 3.67 \\
 Llama 2 70$b^q$  & \cmark & \xmark & 3.63 \\
 \texttt{gpt-3.5-turbo} & \cmark & \xmark & 4.34 \\

  \bottomrule

\end{tabular}
}
\caption{Ablation study on the impact of removing the optimized system prompt and in-context learning (ICL) examples optimized using our method. In the absence of the optimized system prompt, a basic system prompt is provided. Our method consistently outperforms all ablation variants across all models.}
\label{tab:ablation_icl_prompt}
\vspace{-20pt}
\end{center}
\end{table}

\noindent \textbf{Ablation on search algorithms}. 
Table \ref{tab:ablation_search_algo} presents the effect of search algorithms on prompt optimization. We have kept the state and action definitions the same and have only changed the underlying search algorithm.  In this experiment, we ensured that MC and Beam sample the same number of prompts, i.e., same cost, whereas greedy search has a lower cost because the beam width is fixed at 1. More implementation details can be found in Appendix \ref{sec:impl_details}. \ours with beam search gives the best results, depicting the need for thoughtful search and efficient optimization for optimal results.

\begin{table}[!t]
\begin{center}

\begin{tabular}{ c c c  } 
    \toprule
    \textbf{Model} & \textbf{Search} & \textbf{Avg.} \\
    
    \midrule
    Mistral 7b (Instruct) & Beam  & \textbf{4.06} \\
    \midrule
    
    Mistral 7b (Instruct) &  MC  & 4.02 \\
    Mistral 7b (Instruct) & Greedy & 4.02 \\
    
    \bottomrule
    
\end{tabular}

\caption{Ablation study on search methods. MC: Monte Carlo Search; Greedy: greedy search; Beam: beam search. Our method outperforms all other search algorithms tested in the ablation study.}
\vspace{-10pt}
\label{tab:ablation_search_algo}
\end{center}
\end{table}


\begin{table}[!t]
\begin{center}
\resizebox{0.9\linewidth}{!}{%
\begin{tabular}{ c c c c } 
    \toprule
    \multirow{3}{*}{\textbf{Model}} & \textbf{Dynamic} &  \textbf{Dynamic}  &\multirow{3}{*}{\textbf{Avg.}} \\
    & \textbf{Reward} &  \textbf{Reward } \\
    & \textbf{Prompt} & \textbf{ICL} \\
    
    \midrule
    Mistral 7b (Instruct) & \cmark  & \cmark & \textbf{4.06} \\
    \midrule 
    
    Mistral 7b (Instruct) &  \xmark  & \cmark  & 4.02 \\
    Mistral 7b (Instruct) & \cmark & \xmark & 3.86 \\

    \bottomrule
    
\end{tabular}}

\caption{Ablation study on dynamic rewarding, examining its removal from system prompt and ICL example optimization. Our method, utilizing dynamic rewarding for both prompts and ICL examples, consistently outperforms both ablation variants.}
\label{tab:ablation_method}
\end{center}
\vspace{-20pt}
\end{table}

\noindent \textbf{Ablation on dynamic rewarding}.
We performed ablations on the dynamic rewarding mechanism. Table \ref{tab:ablation_method} depicts that \ours, with its current setting of using dynamic rewards for system prompt and ICL optimization, works the best. The in-context examples and prompts without using Dynamic rewarding are also optimized by `static rewarding' for a fair comparison, i.e., we ask the Optimizer to optimize all the rewards all the time. More details can be found in Appendix \ref{sec:impl_details}.

\begin{figure}[!t]
    \centering
    \includegraphics[ width=\linewidth]{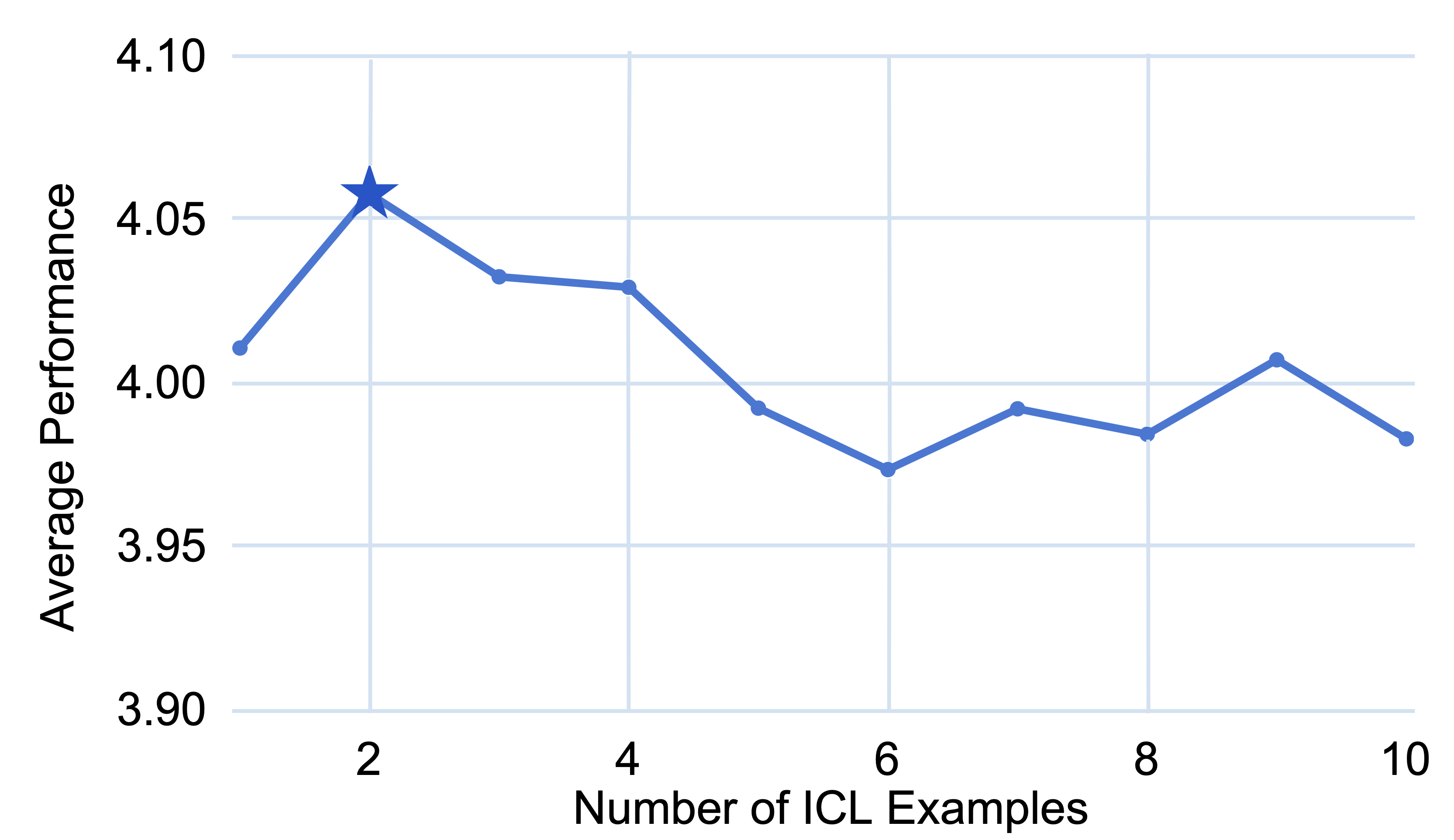}
    \caption{Performance of Mistral 7b (Instruct)  on varying the number of ICL examples. Two examples give us the best performance with a lower context length cost.}
    \label{fig:icl_variation_chart}
    \vspace{-5pt}
\end{figure}

\noindent \textbf{Effect of the number of in-context examples}.
Figure \ref{fig:icl_variation_chart} visualizes the effect of changing the number of in-context learning examples on alignment performance. The choice of $K = 2$ resulted in the best overall performance for Mistral 7b, ensuring strong alignment at a lower context length cost. Also, as observed in Figure \ref{fig:icl_variation_chart}, higher $K$ does not necessarily improve performance, hinting that the quality of ICL examples is more important. The importance of quality is also highlighted in Table \ref{tab:main_table}, where \ours outperforms URIAL at a lower $K$.

\newcommand{\reduline}[1]{{\color{red}\underline{{\color{black}#1}}}}

\newcommand\dunderline[2][.2pt]{\raisebox{-#1}{\underline{\raisebox{#1}{\smash{\underline{#2}}}}}}

\begin{table}[!t]

\definecolor{Gray}{gray}{0.90}
\newcolumntype{a}{>{\columncolor{Gray}}c}
\centering
\resizebox{1\linewidth}{!}{%
\begin{tabular}{@{}p{10cm}@{}}
\toprule
\textbf{Optimized Alignment Prompt} \\
\midrule
As a helpful and ethical assistant, your primary goal is to provide responses that are accurate, engaging, clear, and emotionally resonant across a wide range of queries. \\
- \ctext[RGB]{230,246,255}{Strive to make complex topics understandable and emotionally engaging, communicating in a human-like and relatable manner. Organize your responses to enhance readability and emotional connection, avoiding overly technical jargon.}  \\
- \ctext[RGB]{233,252,232}{Always acknowledge the limitations of your knowledge, especially when speculating about historical 'what-ifs', future predictions, or interpreting emotions.} \\
- \ctext[RGB]{255,225,255}{Aim for a balance between detailed, informative content and a conversational, engaging tone. Incorporate storytelling elements, examples, analogies, and direct questions to make information relatable.} \\
- \ctext[RGB]{230,246,255}{Avoid overwhelming the user with excessive information; structure your responses to be clear, well-organized, and mindful of the user's cognitive load.}
 \\

 \bottomrule
\end{tabular}%
}
 
\caption{Snippets from the system prompt optimized for \texttt{gpt-3.5-turbo}. The optimized prompt clearly demonstrates improved alignment, addressing potential weaknesses in the model.}
\label{tab:gpt_prompt}
\vspace{-15pt}
\end{table}
\noindent \textbf{Qualitative analysis of optimized prompts}. 
We finally present qualitative results to show \ours' ability to identify a model's alignment weaknesses and tailor system prompts to address them, as shown in Table \ref{tab:gpt_prompt} for \texttt{gpt-3.5-turbo}. The color-coded text in the table highlights specific weaknesses of \texttt{gpt-3.5-turbo} identified by \ours, along with actionable insights. Notably, it highlights \ctext[RGB]{233,252,232}{knowledge limitations of the model}, \ctext[RGB]{255,225,255}{tips to improve engagement} and \ctext[RGB]{230,246,255}{technical verbiage}. For a weaker model like Mistral 7b, \ours identifies the problem of repetitive tokens, which is absent in a strong model like \texttt{gpt-3.5-turbo}. Complete optimized prompts for both models, along with detailed annotations on the differences, can be found in Appendix \ref{sec:prompt_case_study}.  


\vspace{-5pt}
\section{Conclusion}
\vspace{-5pt}


This paper introduced Dynamic Rewarding with Prompt Optimization (\ours), a tuning-free approach for self-aligning LLMs. \ours integrates a novel dynamic rewarding mechanism into a search-based prompt optimization framework, enabling LLMs to self-improve model-specific alignment weaknesses adaptively. Experiments on eight LLMs show that \ours-enhanced base models outperform SFT/RLHF-tuned counterparts, and its optimized prompts surpass those by human experts. \ours's adaptability and efficiency offer a promising path toward more personalized AI systems.

\newpage

\section*{Limitations}

While \ours demonstrates significant advancements in tuning-free self-alignment of LLMs, there are a few potential limitations to discuss.

\noindent \textbf{Optimization cost.}
Tuning-free alignment does not come as a free lunch. Ideally, optimizing the alignment prompt for each query would probably be more effective, but its computational overhead is prohibitive. This concern is similar to the decoding-based alignment, where alignment-guided decoding needs to run per query. However, \ours requires only a one-time optimization for each LLM, allowing the optimized alignment prompt to be stored in the LLM memory for future use, significantly reducing the overhead. A detailed analysis of the cost of \ours can be found at \ref{sec:i_cost}.

\noindent \textbf{Computational overhead.}
Compared to SFT / RLHF-tuned models, the increase of input context for the optimized and complex prompt in \ours induces a marginal computational overhead. With advancements in modern LLMs, such as larger context windows, we believe this computational overhead is manageable. Moreover, once an optimized prompt is available with \ours, prompt compression techniques can further reduce the prompt length without sacrificing the performance, which future works can explore.

\noindent \textbf{Automatic rewarding}.
Another potential limitation we noticed is the potential oversight of the internal rewarding process in \ours, which is fully automatic. For example, imprecise rewards might be assigned by dynamic rewarding, leading to undesirable behaviors. We acknowledge this potential issue and have manually reviewed the optimized prompt, finding no severe issues associated with this automatic optimization process. Future work should develop systematic methods to monitor and ensure the accuracy of the reward assignments and the resulting model behaviors. 

\noindent \textbf{Self-correction ability of LLMs}. 
The self-correction ability of LLMs may also be a potential limitation.
When optimizing the system prompt and in-context examples, we rely on LLM-generated feedback, which may occasionally be inaccurate. Upon analyzing feedback traces, we observed that while some feedback was overly critical, it was predominantly constructive. Importantly, the search process mitigates the impact of such overly critical or incorrect feedback on the overall optimization quality. Future work may explore additional guardrails to further ensure the correctness and reliability of LLM-generated feedback throughout the process.

\noindent \textbf{Combination with fine-tuning.}
One may naturally wonder whether \ours can be used to synthesize alignment data and combined with fine-tuning methods to further boost the alignment performance. The answer is yes; however, as highlighted in the paper, one of \ours's unique advantages is its adaptivity, allowing quick adaptation to a new set of reward or user-specific requirements. We value such property and leave the combination of \ours with fine-tuning for future works. 

\noindent \textbf{Capacity assumptions of models.} 
There are certain assumptions on the models involved in \ours. First of all, \ours leverages a strong LLM, specifically GPT-4, as the optimizer to maximize the performance of dynamic rewarding and alignment feedback. Future research could explore other optimizer models, including open-source options, to democratize the application of \ours. Additionally, \ours imposes certain capacity requirements on the base models. Given the complexity of our optimized alignment prompt, smaller and less powerful LLMs, such as LLaMA-7b~\cite{touvron2023llama}, may not experience dramatic improvements through \ours, although some enhancement is still possible. Our assumption is that better pre-trained and instruction-following models have greater potential to be augmented by \ours. We leave such a meaningful question to future research, studying the alignment potential and threshold of LLMs.

Finally, future work may explore further enhancements to the dynamic rewarding mechanism and broader applications of \ours across different domains and tasks.

\section*{Acknowledgment}

We thank the anonymous reviewers for their constructive comments and suggestions. We are also grateful to Enze Ma for integrating \ours into \texttt{LLM Reasoners} and for the valuable discussions with members of MixLab. This work was supported by the OpenAI Agentic AI Research Grant Program. The views and conclusions expressed in this paper are those of the authors and do not necessarily reflect the views of the funding agencies.



\bibliography{main}

\begin{thebibliography}{46}
\providecommand{\natexlab}[1]{#1}

\bibitem[{Achiam et~al.(2023)Achiam, Adler, Agarwal, Ahmad, Akkaya, Aleman, Almeida, Altenschmidt, Altman, Anadkat et~al.}]{achiam2023gpt}
Josh Achiam, Steven Adler, Sandhini Agarwal, Lama Ahmad, Ilge Akkaya, Florencia~Leoni Aleman, Diogo Almeida, Janko Altenschmidt, Sam Altman, Shyamal Anadkat, et~al. 2023.
\newblock Gpt-4 technical report.
\newblock \emph{arXiv preprint arXiv:2303.08774}.

\bibitem[{AI@Meta(2024)}]{llama3modelcard}
AI@Meta. 2024.
\newblock \href {https://github.com/meta-llama/llama3/blob/main/MODEL_CARD.md} {Llama 3 model card}.

\bibitem[{Bai et~al.(2022{\natexlab{a}})Bai, Jones, Ndousse, Askell, Chen, DasSarma, Drain, Fort, Ganguli, Henighan et~al.}]{bai2022training}
Yuntao Bai, Andy Jones, Kamal Ndousse, Amanda Askell, Anna Chen, Nova DasSarma, Dawn Drain, Stanislav Fort, Deep Ganguli, Tom Henighan, et~al. 2022{\natexlab{a}}.
\newblock Training a helpful and harmless assistant with reinforcement learning from human feedback.
\newblock \emph{arXiv preprint arXiv:2204.05862}.

\bibitem[{Bai et~al.(2022{\natexlab{b}})Bai, Kadavath, Kundu, Askell, Kernion, Jones, Chen, Goldie, Mirhoseini, McKinnon et~al.}]{bai2022constitutional}
Yuntao Bai, Saurav Kadavath, Sandipan Kundu, Amanda Askell, Jackson Kernion, Andy Jones, Anna Chen, Anna Goldie, Azalia Mirhoseini, Cameron McKinnon, et~al. 2022{\natexlab{b}}.
\newblock Constitutional ai: Harmlessness from ai feedback.
\newblock \emph{arXiv preprint arXiv:2212.08073}.

\bibitem[{Brown et~al.(2020)Brown, Mann, Ryder, Subbiah, Kaplan, Dhariwal, Neelakantan, Shyam, Sastry, Askell et~al.}]{brown2020language}
Tom Brown, Benjamin Mann, Nick Ryder, Melanie Subbiah, Jared~D Kaplan, Prafulla Dhariwal, Arvind Neelakantan, Pranav Shyam, Girish Sastry, Amanda Askell, et~al. 2020.
\newblock Language models are few-shot learners.
\newblock \emph{Advances in neural information processing systems}, 33:1877--1901.

\bibitem[{Burns et~al.(2023)Burns, Izmailov, Kirchner, Baker, Gao, Aschenbrenner, Chen, Ecoffet, Joglekar, Leike et~al.}]{burns2023weak}
Collin Burns, Pavel Izmailov, Jan~Hendrik Kirchner, Bowen Baker, Leo Gao, Leopold Aschenbrenner, Yining Chen, Adrien Ecoffet, Manas Joglekar, Jan Leike, et~al. 2023.
\newblock Weak-to-strong generalization: Eliciting strong capabilities with weak supervision.
\newblock \emph{arXiv preprint arXiv:2312.09390}.

\bibitem[{Cao et~al.(2024)Cao, Lu, Lu, Chen, Ren, Xiang, Liu, Lu, He, Han et~al.}]{cao2024towards}
Boxi Cao, Keming Lu, Xinyu Lu, Jiawei Chen, Mengjie Ren, Hao Xiang, Peilin Liu, Yaojie Lu, Ben He, Xianpei Han, et~al. 2024.
\newblock Towards scalable automated alignment of llms: A survey.
\newblock \emph{arXiv preprint arXiv:2406.01252}.

\bibitem[{Chowdhery et~al.(2023)Chowdhery, Narang, Devlin, Bosma, Mishra, Roberts, Barham, Chung, Sutton, Gehrmann et~al.}]{chowdhery2023palm}
Aakanksha Chowdhery, Sharan Narang, Jacob Devlin, Maarten Bosma, Gaurav Mishra, Adam Roberts, Paul Barham, Hyung~Won Chung, Charles Sutton, Sebastian Gehrmann, et~al. 2023.
\newblock Palm: Scaling language modeling with pathways.
\newblock \emph{Journal of Machine Learning Research}, 24(240):1--113.

\bibitem[{Dong et~al.(2022)Dong, Li, Dai, Zheng, Wu, Chang, Sun, Xu, and Sui}]{dong2022survey}
Qingxiu Dong, Lei Li, Damai Dai, Ce~Zheng, Zhiyong Wu, Baobao Chang, Xu~Sun, Jingjing Xu, and Zhifang Sui. 2022.
\newblock A survey on in-context learning.
\newblock \emph{arXiv preprint arXiv:2301.00234}.

\bibitem[{Fernando et~al.(2023)Fernando, Banarse, Michalewski, Osindero, and Rockt{\"a}schel}]{fernando2023promptbreeder}
Chrisantha Fernando, Dylan Banarse, Henryk Michalewski, Simon Osindero, and Tim Rockt{\"a}schel. 2023.
\newblock Promptbreeder: Self-referential self-improvement via prompt evolution.
\newblock \emph{arXiv preprint arXiv:2309.16797}.

\bibitem[{Ganguli et~al.(2022)Ganguli, Lovitt, Kernion, Askell, Bai, Kadavath, Mann, Perez, Schiefer, Ndousse, Jones, Bowman, Chen, Conerly, Dassarma, Drain, Elhage, El-Showk, Fort, Dodds, Henighan, Hernandez, Hume, Jacobson, Johnston, Kravec, Olsson, Ringer, Tran-Johnson, Amodei, Brown, Joseph, McCandlish, Olah, Kaplan, and Clark}]{Ganguli2022RedTL}
Deep Ganguli, Liane Lovitt, John Kernion, Amanda Askell, Yuntao Bai, Saurav Kadavath, Benjamin Mann, Ethan Perez, Nicholas Schiefer, Kamal Ndousse, Andy Jones, Sam Bowman, Anna Chen, Tom Conerly, Nova Dassarma, Dawn Drain, Nelson Elhage, Sheer El-Showk, Stanislav Fort, Zachary Dodds, Tom Henighan, Danny Hernandez, Tristan Hume, Josh Jacobson, Scott Johnston, Shauna Kravec, Catherine Olsson, Sam Ringer, Eli Tran-Johnson, Dario Amodei, Tom~B. Brown, Nicholas Joseph, Sam McCandlish, Christopher Olah, Jared Kaplan, and Jack Clark. 2022.
\newblock \href {https://api.semanticscholar.org/CorpusID:252355458} {Red teaming language models to reduce harms: Methods, scaling behaviors, and lessons learned}.
\newblock \emph{ArXiv}, abs/2209.07858.

\bibitem[{Guo et~al.(2024)Guo, Yao, Shen, Wei, Zhang, Wang, and Liu}]{guo2024human}
Hongyi Guo, Yuanshun Yao, Wei Shen, Jiaheng Wei, Xiaoying Zhang, Zhaoran Wang, and Yang Liu. 2024.
\newblock Human-instruction-free llm self-alignment with limited samples.
\newblock \emph{arXiv preprint arXiv:2401.06785}.

\bibitem[{Han(2023)}]{han2023context}
Xiaochuang Han. 2023.
\newblock In-context alignment: Chat with vanilla language models before fine-tuning.
\newblock \emph{arXiv preprint arXiv:2308.04275}.

\bibitem[{Hao et~al.(2024)Hao, Gu, Luo, Liu, Shao, Wang, Xie, Ma, Samavedhi, Gao, Wang, and Hu}]{hao2024llm}
Shibo Hao, Yi~Gu, Haotian Luo, Tianyang Liu, Xiyan Shao, Xinyuan Wang, Shuhua Xie, Haodi Ma, Adithya Samavedhi, Qiyue Gao, Zhen Wang, and Zhiting Hu. 2024.
\newblock \href {https://arxiv.org/abs/2404.05221} {Llm reasoners: New evaluation, library, and analysis of step-by-step reasoning with large language models}.
\newblock \emph{Preprint}, arXiv:2404.05221.

\bibitem[{Hao et~al.(2023)Hao, Gu, Ma, Hong, Wang, Wang, and Hu}]{hao2023reasoning}
Shibo Hao, Yi~Gu, Haodi Ma, Joshua Hong, Zhen Wang, Daisy Wang, and Zhiting Hu. 2023.
\newblock Reasoning with language model is planning with world model.
\newblock In \emph{Proceedings of the 2023 Conference on Empirical Methods in Natural Language Processing}, pages 8154--8173.

\bibitem[{Huang et~al.(2024)Huang, Sengupta, Bonadiman, Lai, Gupta, Pappas, Mansour, Kirchoff, and Roth}]{huang2024deal}
James~Y Huang, Sailik Sengupta, Daniele Bonadiman, Yi-an Lai, Arshit Gupta, Nikolaos Pappas, Saab Mansour, Katrin Kirchoff, and Dan Roth. 2024.
\newblock Deal: Decoding-time alignment for large language models.
\newblock \emph{arXiv preprint arXiv:2402.06147}.

\bibitem[{Jiang et~al.(2023)Jiang, Sablayrolles, Mensch, Bamford, Chaplot, de~Las~Casas, Bressand, Lengyel, Lample, Saulnier, Lavaud, Lachaux, Stock, Scao, Lavril, Wang, Lacroix, and Sayed}]{Jiang2023Mistral7}
Albert~Qiaochu Jiang, Alexandre Sablayrolles, Arthur Mensch, Chris Bamford, Devendra~Singh Chaplot, Diego de~Las~Casas, Florian Bressand, Gianna Lengyel, Guillaume Lample, Lucile Saulnier, L'elio~Renard Lavaud, Marie-Anne Lachaux, Pierre Stock, Teven~Le Scao, Thibaut Lavril, Thomas Wang, Timoth{\'e}e Lacroix, and William~El Sayed. 2023.
\newblock \href {https://api.semanticscholar.org/CorpusID:263830494} {Mistral 7b}.
\newblock \emph{ArXiv}, abs/2310.06825.

\bibitem[{Khanov et~al.(2024)Khanov, Burapacheep, and Li}]{khanov2024args}
Maxim Khanov, Jirayu Burapacheep, and Yixuan Li. 2024.
\newblock Args: Alignment as reward-guided search.
\newblock \emph{arXiv preprint arXiv:2402.01694}.

\bibitem[{Kim et~al.(2023)Kim, Bae, Shin, Kang, Kwak, Yoo, and Seo}]{kim2023aligning}
Sungdong Kim, Sanghwan Bae, Jamin Shin, Soyoung Kang, Donghyun Kwak, Kang~Min Yoo, and Minjoon Seo. 2023.
\newblock Aligning large language models through synthetic feedback.
\newblock \emph{arXiv preprint arXiv:2305.13735}.

\bibitem[{Kong et~al.(2024)Kong, Wang, Mu, Du, Zhuang, Zhou, Song, Zhang, Wang, and Zhang}]{kong2024aligning}
Lingkai Kong, Haorui Wang, Wenhao Mu, Yuanqi Du, Yuchen Zhuang, Yifei Zhou, Yue Song, Rongzhi Zhang, Kai Wang, and Chao Zhang. 2024.
\newblock Aligning large language models with representation editing: A control perspective.
\newblock \emph{arXiv preprint arXiv:2406.05954}.

\bibitem[{Lee et~al.(2023)Lee, Phatale, Mansoor, Lu, Mesnard, Bishop, Carbune, and Rastogi}]{lee2023rlaif}
Harrison Lee, Samrat Phatale, Hassan Mansoor, Kellie Lu, Thomas Mesnard, Colton Bishop, Victor Carbune, and Abhinav Rastogi. 2023.
\newblock Rlaif: Scaling reinforcement learning from human feedback with ai feedback.
\newblock \emph{arXiv preprint arXiv:2309.00267}.

\bibitem[{Li et~al.(2024)Li, Patel, Vi{\'e}gas, Pfister, and Wattenberg}]{li2024inference}
Kenneth Li, Oam Patel, Fernanda Vi{\'e}gas, Hanspeter Pfister, and Martin Wattenberg. 2024.
\newblock Inference-time intervention: Eliciting truthful answers from a language model.
\newblock \emph{Advances in Neural Information Processing Systems}, 36.

\bibitem[{Li et~al.(2023{\natexlab{a}})Li, Yu, Zhou, Schick, Zettlemoyer, Levy, Weston, and Lewis}]{li2023self}
Xian Li, Ping Yu, Chunting Zhou, Timo Schick, Luke Zettlemoyer, Omer Levy, Jason Weston, and Mike Lewis. 2023{\natexlab{a}}.
\newblock Self-alignment with instruction backtranslation.
\newblock \emph{arXiv preprint arXiv:2308.06259}.

\bibitem[{Li et~al.(2023{\natexlab{b}})Li, Zhang, Dubois, Taori, Gulrajani, Guestrin, Liang, and Hashimoto}]{alpaca_eval}
Xuechen Li, Tianyi Zhang, Yann Dubois, Rohan Taori, Ishaan Gulrajani, Carlos Guestrin, Percy Liang, and Tatsunori~B. Hashimoto. 2023{\natexlab{b}}.
\newblock Alpacaeval: An automatic evaluator of instruction-following models.
\newblock \url{https://github.com/tatsu-lab/alpaca_eval}.

\bibitem[{Li et~al.(2023{\natexlab{c}})Li, Wei, Zhao, Zhang, and Zhang}]{li2023rain}
Yuhui Li, Fangyun Wei, Jinjing Zhao, Chao Zhang, and Hongyang Zhang. 2023{\natexlab{c}}.
\newblock Rain: Your language models can align themselves without finetuning.
\newblock \emph{arXiv preprint arXiv:2309.07124}.

\bibitem[{Lin et~al.(2024{\natexlab{a}})Lin, Ravichander, Lu, Dziri, Sclar, Chandu, Bhagavatula, and Choi}]{Lin2024ReAlign}
Bill~Yuchen Lin, Abhilasha Ravichander, Ximing Lu, Nouha Dziri, Melanie Sclar, Khyathi Chandu, Chandra Bhagavatula, and Yejin Choi. 2024{\natexlab{a}}.
\newblock \href {https://arxiv.org/abs/2312.01552} {The unlocking spell on base llms: Rethinking alignment via in-context learning}.
\newblock In \emph{International Conference on Learning Representations}.

\bibitem[{Lin et~al.(2024{\natexlab{b}})Lin, Tang, Tang, Yang, Chen, Wang, Xiao, Dang, Gan, and Han}]{lin2023awq}
Ji~Lin, Jiaming Tang, Haotian Tang, Shang Yang, Wei-Ming Chen, Wei-Chen Wang, Guangxuan Xiao, Xingyu Dang, Chuang Gan, and Song Han. 2024{\natexlab{b}}.
\newblock Awq: Activation-aware weight quantization for llm compression and acceleration.
\newblock In \emph{MLSys}.

\bibitem[{Madaan et~al.(2024)Madaan, Tandon, Gupta, Hallinan, Gao, Wiegreffe, Alon, Dziri, Prabhumoye, Yang et~al.}]{madaan2024self}
Aman Madaan, Niket Tandon, Prakhar Gupta, Skyler Hallinan, Luyu Gao, Sarah Wiegreffe, Uri Alon, Nouha Dziri, Shrimai Prabhumoye, Yiming Yang, et~al. 2024.
\newblock Self-refine: Iterative refinement with self-feedback.
\newblock \emph{Advances in Neural Information Processing Systems}, 36.

\bibitem[{Ouyang et~al.(2022)Ouyang, Wu, Jiang, Almeida, Wainwright, Mishkin, Zhang, Agarwal, Slama, Ray, Schulman, Hilton, Kelton, Miller, Simens, Askell, Welinder, Christiano, Leike, and Lowe}]{ouyang2022training}
Long Ouyang, Jeff Wu, Xu~Jiang, Diogo Almeida, Carroll~L. Wainwright, Pamela Mishkin, Chong Zhang, Sandhini Agarwal, Katarina Slama, Alex Ray, John Schulman, Jacob Hilton, Fraser Kelton, Luke Miller, Maddie Simens, Amanda Askell, Peter Welinder, Paul Christiano, Jan Leike, and Ryan Lowe. 2022.
\newblock \href {https://arxiv.org/abs/2203.02155} {Training language models to follow instructions with human feedback}.
\newblock \emph{Preprint}, arXiv:2203.02155.

\bibitem[{Pryzant et~al.(2023)Pryzant, Iter, Li, Lee, Zhu, and Zeng}]{pryzant2023automatic}
Reid Pryzant, Dan Iter, Jerry Li, Yin~Tat Lee, Chenguang Zhu, and Michael Zeng. 2023.
\newblock Automatic prompt optimization with" gradient descent" and beam search.
\newblock \emph{arXiv preprint arXiv:2305.03495}.

\bibitem[{Reimers and Gurevych(2019)}]{reimers-2019-sentence-bert}
Nils Reimers and Iryna Gurevych. 2019.
\newblock \href {https://arxiv.org/abs/1908.10084} {Sentence-bert: Sentence embeddings using siamese bert-networks}.
\newblock In \emph{Proceedings of the 2019 Conference on Empirical Methods in Natural Language Processing}. Association for Computational Linguistics.

\bibitem[{Rubin et~al.(2021)Rubin, Herzig, and Berant}]{rubin2021learning}
Ohad Rubin, Jonathan Herzig, and Jonathan Berant. 2021.
\newblock Learning to retrieve prompts for in-context learning.
\newblock \emph{arXiv preprint arXiv:2112.08633}.

\bibitem[{Sun et~al.(2024)Sun, Shen, Zhou, Zhang, Chen, Cox, Yang, and Gan}]{sun2024principle}
Zhiqing Sun, Yikang Shen, Qinhong Zhou, Hongxin Zhang, Zhenfang Chen, David Cox, Yiming Yang, and Chuang Gan. 2024.
\newblock Principle-driven self-alignment of language models from scratch with minimal human supervision.
\newblock \emph{Advances in Neural Information Processing Systems}, 36.

\bibitem[{Touvron et~al.(2023{\natexlab{a}})Touvron, Lavril, Izacard, Martinet, Lachaux, Lacroix, Rozi{\`e}re, Goyal, Hambro, Azhar et~al.}]{touvron2023llama}
Hugo Touvron, Thibaut Lavril, Gautier Izacard, Xavier Martinet, Marie-Anne Lachaux, Timoth{\'e}e Lacroix, Baptiste Rozi{\`e}re, Naman Goyal, Eric Hambro, Faisal Azhar, et~al. 2023{\natexlab{a}}.
\newblock Llama: Open and efficient foundation language models.
\newblock \emph{arXiv preprint arXiv:2302.13971}.

\bibitem[{Touvron et~al.(2023{\natexlab{b}})Touvron, Martin, Stone, Albert, Almahairi, Babaei, Bashlykov, Batra, Bhargava, Bhosale, Bikel, Blecher, Ferrer, Chen, Cucurull, Esiobu, Fernandes, Fu, Fu, Fuller, Gao, Goswami, Goyal, Hartshorn, Hosseini, Hou, Inan, Kardas, Kerkez, Khabsa, Kloumann, Korenev, Koura, Lachaux, Lavril, Lee, Liskovich, Lu, Mao, Martinet, Mihaylov, Mishra, Molybog, Nie, Poulton, Reizenstein, Rungta, Saladi, Schelten, Silva, Smith, Subramanian, Tan, Tang, Taylor, Williams, Kuan, Xu, Yan, Zarov, Zhang, Fan, Kambadur, Narang, Rodriguez, Stojnic, Edunov, and Scialom}]{Touvron2023Llama2O}
Hugo Touvron, Louis Martin, Kevin~R. Stone, Peter Albert, Amjad Almahairi, Yasmine Babaei, Nikolay Bashlykov, Soumya Batra, Prajjwal Bhargava, Shruti Bhosale, Daniel~M. Bikel, Lukas Blecher, Cristian~Cant{\'o}n Ferrer, Moya Chen, Guillem Cucurull, David Esiobu, Jude Fernandes, Jeremy Fu, Wenyin Fu, Brian Fuller, Cynthia Gao, Vedanuj Goswami, Naman Goyal, Anthony~S. Hartshorn, Saghar Hosseini, Rui Hou, Hakan Inan, Marcin Kardas, Viktor Kerkez, Madian Khabsa, Isabel~M. Kloumann, A.~V. Korenev, Punit~Singh Koura, Marie-Anne Lachaux, Thibaut Lavril, Jenya Lee, Diana Liskovich, Yinghai Lu, Yuning Mao, Xavier Martinet, Todor Mihaylov, Pushkar Mishra, Igor Molybog, Yixin Nie, Andrew Poulton, Jeremy Reizenstein, Rashi Rungta, Kalyan Saladi, Alan Schelten, Ruan Silva, Eric~Michael Smith, R.~Subramanian, Xia Tan, Binh Tang, Ross Taylor, Adina Williams, Jian~Xiang Kuan, Puxin Xu, Zhengxu Yan, Iliyan Zarov, Yuchen Zhang, Angela Fan, Melanie Kambadur, Sharan Narang, Aurelien Rodriguez, Robert Stojnic, Sergey Edunov, and
  Thomas Scialom. 2023{\natexlab{b}}.
\newblock \href {https://api.semanticscholar.org/CorpusID:259950998} {Llama 2: Open foundation and fine-tuned chat models}.
\newblock \emph{ArXiv}, abs/2307.09288.

\bibitem[{Wang et~al.(2024{\natexlab{a}})Wang, Ma, Meng, Qin, Shen, Zhang, Wu, Liu, Bian, Xu et~al.}]{wang2024step}
Haoyu Wang, Guozheng Ma, Ziqiao Meng, Zeyu Qin, Li~Shen, Zhong Zhang, Bingzhe Wu, Liu Liu, Yatao Bian, Tingyang Xu, et~al. 2024{\natexlab{a}}.
\newblock Step-on-feet tuning: Scaling self-alignment of llms via bootstrapping.
\newblock \emph{arXiv preprint arXiv:2402.07610}.

\bibitem[{Wang et~al.(2024{\natexlab{b}})Wang, Zhang, Li, Tan, Wang, Ren, Jiang, and Qiu}]{wang2024inferaligner}
Pengyu Wang, Dong Zhang, Linyang Li, Chenkun Tan, Xinghao Wang, Ke~Ren, Botian Jiang, and Xipeng Qiu. 2024{\natexlab{b}}.
\newblock Inferaligner: Inference-time alignment for harmlessness through cross-model guidance.
\newblock \emph{arXiv preprint arXiv:2401.11206}.

\bibitem[{Wang et~al.(2023)Wang, Li, Wang, Bai, Luo, Zhang, Jojic, Xing, and Hu}]{wang2023promptagent}
Xinyuan Wang, Chenxi Li, Zhen Wang, Fan Bai, Haotian Luo, Jiayou Zhang, Nebojsa Jojic, Eric Xing, and Zhiting Hu. 2023.
\newblock Promptagent: Strategic planning with language models enables expert-level prompt optimization.
\newblock In \emph{The Twelfth International Conference on Learning Representations}.

\bibitem[{Wang et~al.(2022)Wang, Kordi, Mishra, Liu, Smith, Khashabi, and Hajishirzi}]{wang2022self}
Yizhong Wang, Yeganeh Kordi, Swaroop Mishra, Alisa Liu, Noah~A Smith, Daniel Khashabi, and Hannaneh Hajishirzi. 2022.
\newblock Self-instruct: Aligning language models with self-generated instructions.
\newblock \emph{arXiv preprint arXiv:2212.10560}.

\bibitem[{Wu et~al.(2024)Wu, Arora, Wang, Geiger, Jurafsky, Manning, and Potts}]{wu2024reft}
Zhengxuan Wu, Aryaman Arora, Zheng Wang, Atticus Geiger, Dan Jurafsky, Christopher~D Manning, and Christopher Potts. 2024.
\newblock Reft: Representation finetuning for language models.
\newblock \emph{arXiv preprint arXiv:2404.03592}.

\bibitem[{Xu et~al.(2022)Xu, Chen, Du, Shao, Wang, Li, and Yang}]{xu2022gps}
Hanwei Xu, Yujun Chen, Yulun Du, Nan Shao, Yanggang Wang, Haiyu Li, and Zhilin Yang. 2022.
\newblock Gps: Genetic prompt search for efficient few-shot learning.
\newblock \emph{arXiv preprint arXiv:2210.17041}.

\bibitem[{Yang et~al.(2023)Yang, Wang, Lu, Liu, Le, Zhou, and Chen}]{yang2023large}
Chengrun Yang, Xuezhi Wang, Yifeng Lu, Hanxiao Liu, Quoc~V Le, Denny Zhou, and Xinyun Chen. 2023.
\newblock Large language models as optimizers.
\newblock \emph{arXiv preprint arXiv:2309.03409}.

\bibitem[{Zhao et~al.(2024)Zhao, Andriushchenko, Croce, and Flammarion}]{zhao2024context}
Hao Zhao, Maksym Andriushchenko, Francesco Croce, and Nicolas Flammarion. 2024.
\newblock Is in-context learning sufficient for instruction following in llms?
\newblock \emph{arXiv preprint arXiv:2405.19874}.

\bibitem[{Zhou et~al.(2024)Zhou, Liu, Xu, Iyer, Sun, Mao, Ma, Efrat, Yu, Yu et~al.}]{zhou2024lima}
Chunting Zhou, Pengfei Liu, Puxin Xu, Srinivasan Iyer, Jiao Sun, Yuning Mao, Xuezhe Ma, Avia Efrat, Ping Yu, Lili Yu, et~al. 2024.
\newblock Lima: Less is more for alignment.
\newblock \emph{Advances in Neural Information Processing Systems}, 36.

\bibitem[{Zhou et~al.(2022)Zhou, Muresanu, Han, Paster, Pitis, Chan, and Ba}]{zhou2022large}
Yongchao Zhou, Andrei~Ioan Muresanu, Ziwen Han, Keiran Paster, Silviu Pitis, Harris Chan, and Jimmy Ba. 2022.
\newblock Large language models are human-level prompt engineers.
\newblock \emph{arXiv preprint arXiv:2211.01910}.

\bibitem[{Zou et~al.(2023)Zou, Phan, Chen, Campbell, Guo, Ren, Pan, Yin, Mazeika, Dombrowski et~al.}]{zou2023representation}
Andy Zou, Long Phan, Sarah Chen, James Campbell, Phillip Guo, Richard Ren, Alexander Pan, Xuwang Yin, Mantas Mazeika, Ann-Kathrin Dombrowski, et~al. 2023.
\newblock Representation engineering: A top-down approach to ai transparency.
\newblock \emph{arXiv preprint arXiv:2310.01405}.

\end{thebibliography}

\appendix

\newpage

\section{More Implementation Details}
\label{sec:impl_details}

\subsection{Hyper-parameters for \ours}
\begin{table}[H]
\begin{center}
\begin{tabular}{ c c c c } 
    \toprule
    \textbf{Experiment} & \bm{$W$} & \bm{$M$} & \bm{$D$} \\
    \midrule
    ICL optimization & 1 & 1 & 5 \\
    System Prompt optimization & 2 & 3 & 20 \\
    \bottomrule
\end{tabular}
\caption{All the hyper-parameters used by \ours during ICL optimization and system prompt optimization.}
\end{center}
\label{tab:hyper_params_beam}
\end{table}

\subsection{Baselines}

\noindent \textbf{Monte Carlo Search}: Monte Carlo search performs directionless 1-step sampling multiple times. The sampling method was kept the same as \ours; we sampled 120 prompts in this method to keep the cost the same as \ours and ensure a fair comparison.

\noindent \textbf{Greedy Search}: Greedy search is the special case of beam search with beam width $W$ fixed as 1, the sampling method, number of action samples per state $M$ was kept the same as \ours but still as the beam width has decreased in this method the overall cost is lower.

\noindent \textbf{Static Rewarding}: In this method, we keep the search algorithm the same as \ours. Instead of choosing dynamic aspects, we always provide a fixed set of aspects to the optimizer and evaluator. The fixed set of aspects was chosen as helpfulness, clarity, factuality, depth, engagement, and safety i.e. the evaluation aspects. This allowed the static rewarding method to perform the best on evaluation metrics and establish a strong baseline. Note that we keep the number of in-context learning examples as 2 while evaluating this baseline.

\subsection{Seed Samples}
Out of the 180 samples in the sampled dataset, $47.8 \%$ of samples comes from \texttt{AlpacaEval}, $28.9 \%$ from LIMA, and the rest from \texttt{HH-RLHF-redteam}. We ensure a fair evaluation by only sampling examples that are not present in the evaluation dataset.

\subsection{Base ICL Examples}
\label{sec:i_base}
Examples in $\mathcal{I}_{base}$ are classified into two groups: ``unethical'', which teaches the model to handle malicious queries, and ``informative'', which teaches the model to present relevant information in an acceptable format. $\mathcal{I}_{base}$, contains an equal number of ``unethical'' queries and ``informative'' queries. 

\subsection{Cost Analysis of \ours}
\label{sec:i_cost}
\noindent \textbf{System Prompt Optimization}. 
Our optimization process leverages a beam search strategy, with the number of sampled prompts being determined by the parameters $W$ (beam width), $M$ (number of action samples per state), and $D$ (beam depth). Specifically, these parameters result in:

    \begin{enumerate}
        \item  $W \times M \times D$ API calls to the optimizer LLM $\mathcal{O}$ for prompt sampling.
        \item $D$ API calls to LLM for reward selection of seed samples.
        \item $W \times M \times D$ calls to base LLM $\mathcal{B}$ for response generation corresponding to each of the sampled prompts.
        \item $W \times M \times D$ API calls to the evaluator LLM $\mathcal{E}$ for sampled prompt evaluation using seed samples.
    \end{enumerate}

Thus, the overall cost ($C_{\text{system}}$), including both API calls and base LLM inferences, for system prompt optimization can be expressed as:

\begin{align*}
    C_{\text{system}} = & \underbrace{W \times M \times D}_{\text{prompt sampling}}
    + \underbrace{D}_{\text{reward selection}} + \\
    & \underbrace{W \times M \times D}_{\text{response generation}}
    + \underbrace{W \times M \times D}_{\text{prompt evaluation}}
\end{align*}
    
Notably, the reward selection cost is incurred only once, as these results are cached and reused across all models. Moreover, the system prompt optimization is also a one-time process for each model; once optimized, the prompts can be reused without incurring additional costs. This approach ensures that the incurred cost is limited and does not scale with the number of subsequent uses.

\noindent \textbf{ICL Optimization}.
Similar to System prompt optimization we can also use beam search for ICL optimization. The cost for optimizing one ICL example is as follows:

    \begin{enumerate}
        \item A single API call to LLM for reward selection of the example.
        \item $W \times M \times D$ API calls to the evaluator LLM to evaluate the ICL example. (amounting to 5 given the hyperparameters)
        \item $W \times M \times D$ API calls to the optimizer LLM, for optimizing the ICL example.
    \end{enumerate}

Thus, the total cost ($C_{\text{ICL}}$) for ICL optimization can be expressed as:

\begin{align*}
    C_{\text{ICL}} = \quad & (\underbrace{1}_{\text{reward selection}} + \underbrace{W \times M \times D}_{\text{evaluation}} + \\
    & \underbrace{W \times M \times D}_{\text{eptimization}} ) \times N
\end{align*}

where $N$ denotes the number of examples we want to optimize.

ICL examples are model-agnostic and can be reused across different models, thus making the optimization cost a one-time expense per example.

\newpage

\section{Categorized Performance}
\label{sec:cat_perf}

\subsection{Mistral 7b}
\begin{figure}[h]
\centering
\begin{subfigure}[b]{.5\textwidth}
  \centering
  \includegraphics[width=0.95\linewidth]{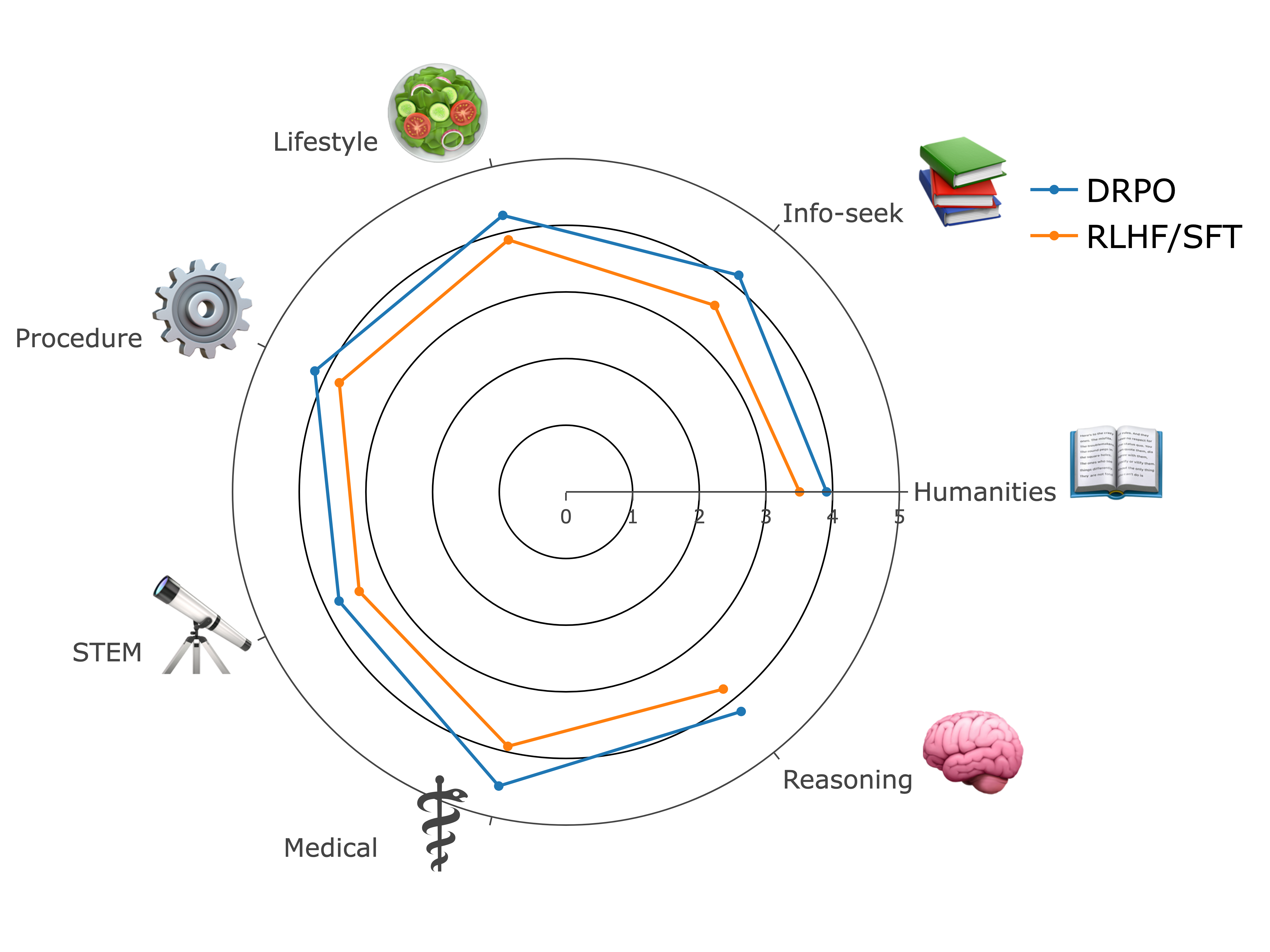}
  \label{fig:cat_mistral_1}
\end{subfigure}%

\vspace{1em}

\begin{subfigure}[b]{.5\textwidth}
  \centering
  \includegraphics[width=0.95\linewidth]{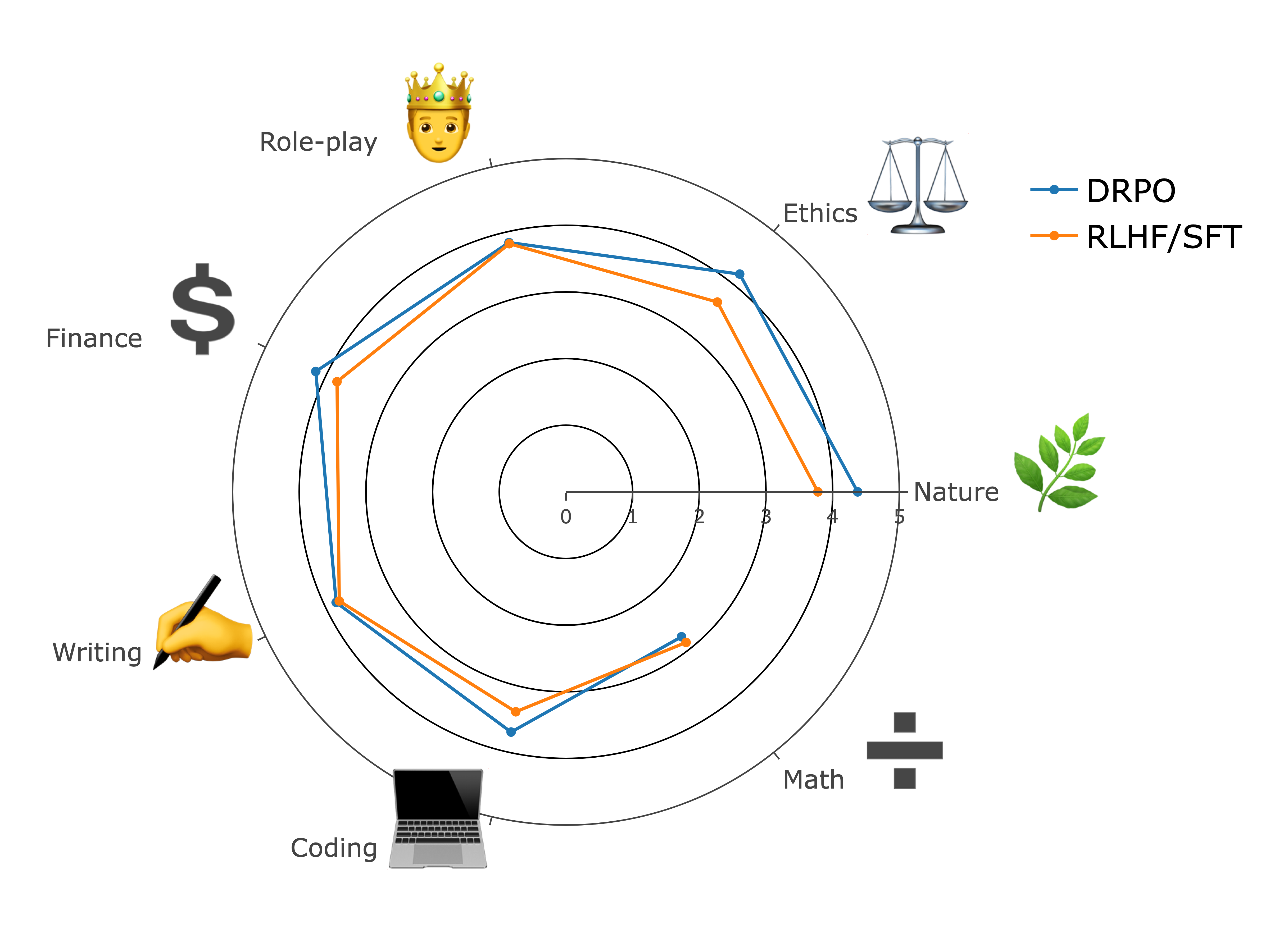}
  \label{fig:cat_mistral_2}
\end{subfigure}
\caption{Categorized performance of Mistral 7b across various domains. Using \ours we see a strong improvement in performance across all domains. Notably, we can see that domains like Humanities, Reasoning, STEM improves significantly. This highlights the fact that base models can benefit a great deal from \ours.  }
\label{fig:categorized_performance_mistral}
\end{figure}

\newpage
\subsection{Llama 2 70b}
\begin{figure}[h]
\centering
\begin{subfigure}[b]{.5\textwidth}
  \centering
  \includegraphics[width=0.95\linewidth]{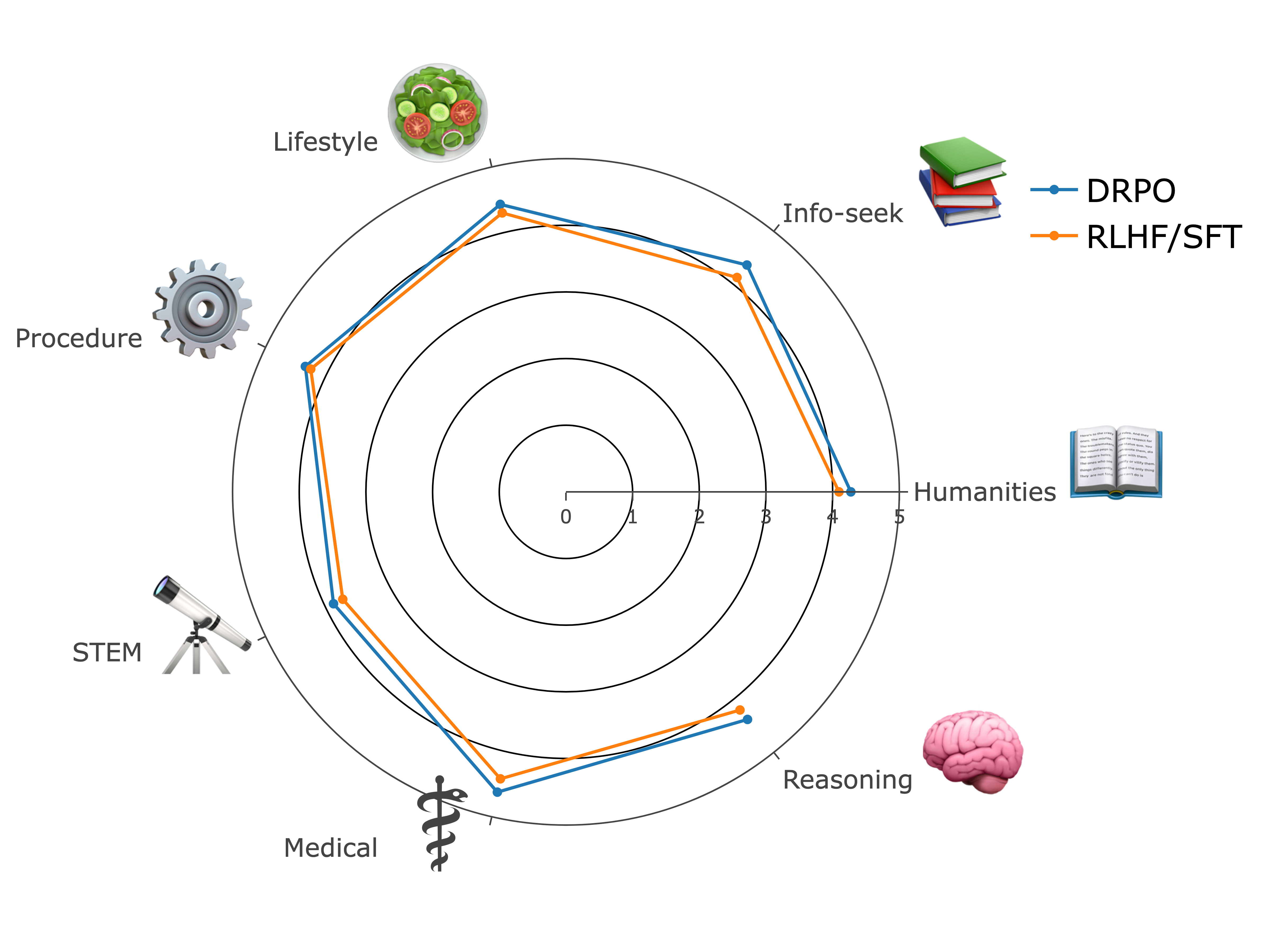}
  \label{fig:cat_llama_1}
\end{subfigure}%

\vspace{1em}

\begin{subfigure}[b]{.5\textwidth}
  \centering
  \includegraphics[width=0.95\linewidth]{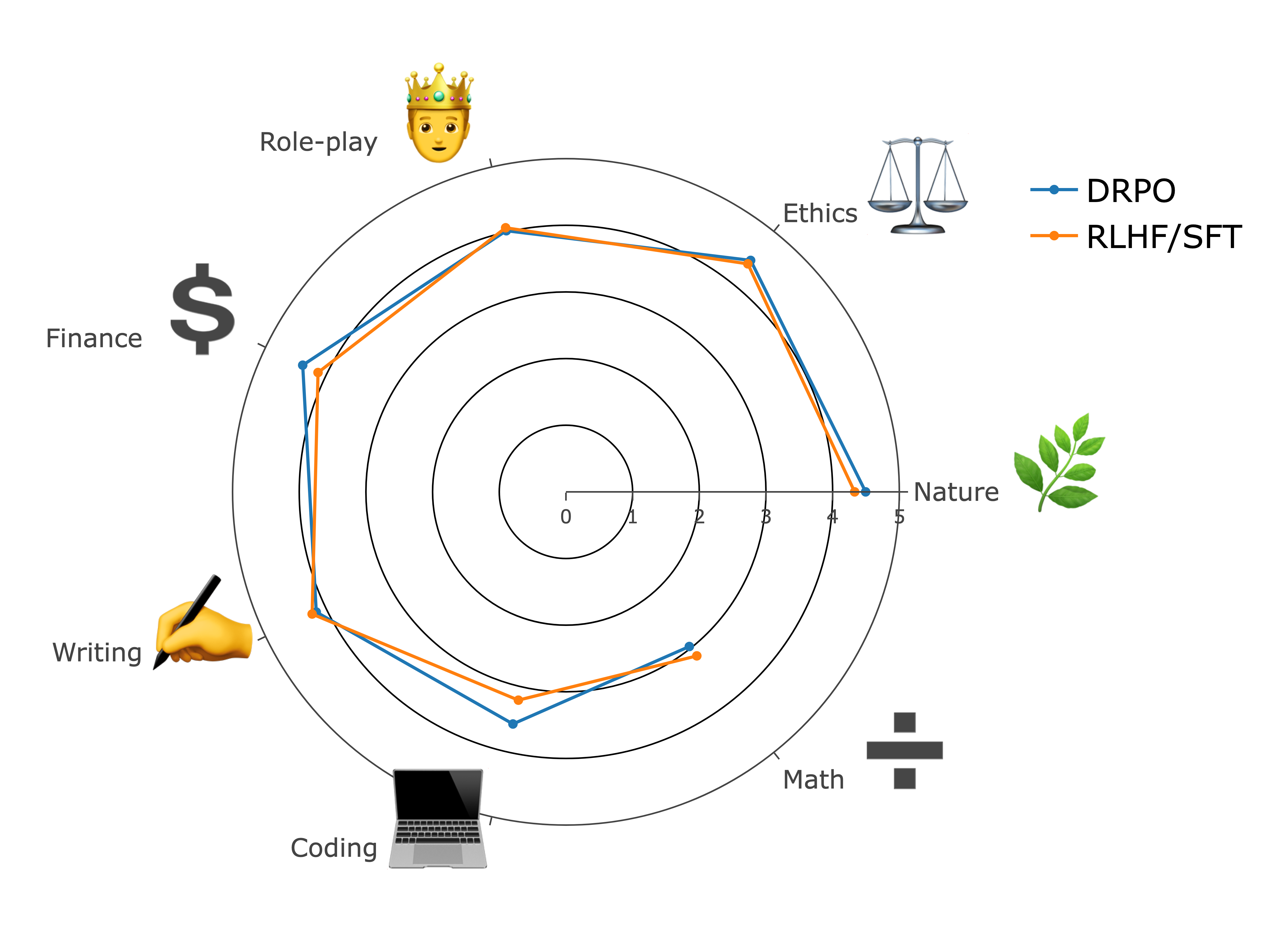}
  \label{fig:cat_llama_2}
\end{subfigure}
\caption{Categorized performance of Llama 2 70$b^q$ across various domains. Using \ours we see an improvement in performance across all domains barring math where we see a small drop. The performance using \ours strongly improves domains such as Info-seek, Coding, and Finance.  }
\label{fig:categorized_performance_llama}
\end{figure}

\newpage
\subsection{\texttt{gpt-3.5-turbo}}

\begin{figure}[h]
\centering
\begin{subfigure}[b]{.5\textwidth}
  \centering
  \includegraphics[width=0.95\linewidth]{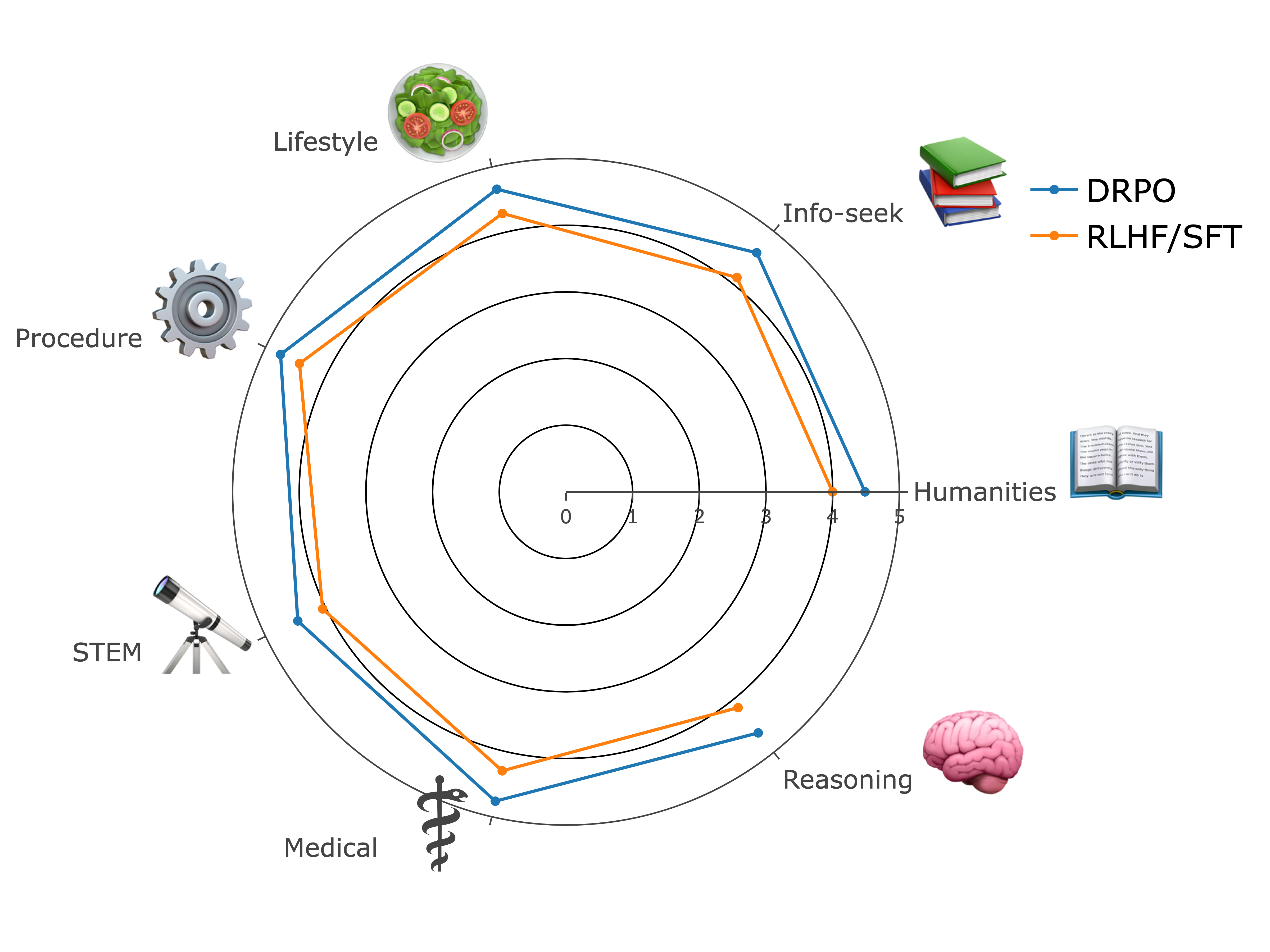}
  \label{fig:cat_gpt_1}
\end{subfigure}%

\vspace{1em}

\begin{subfigure}[b]{.5\textwidth}
  \centering
  \includegraphics[width=0.95\linewidth]{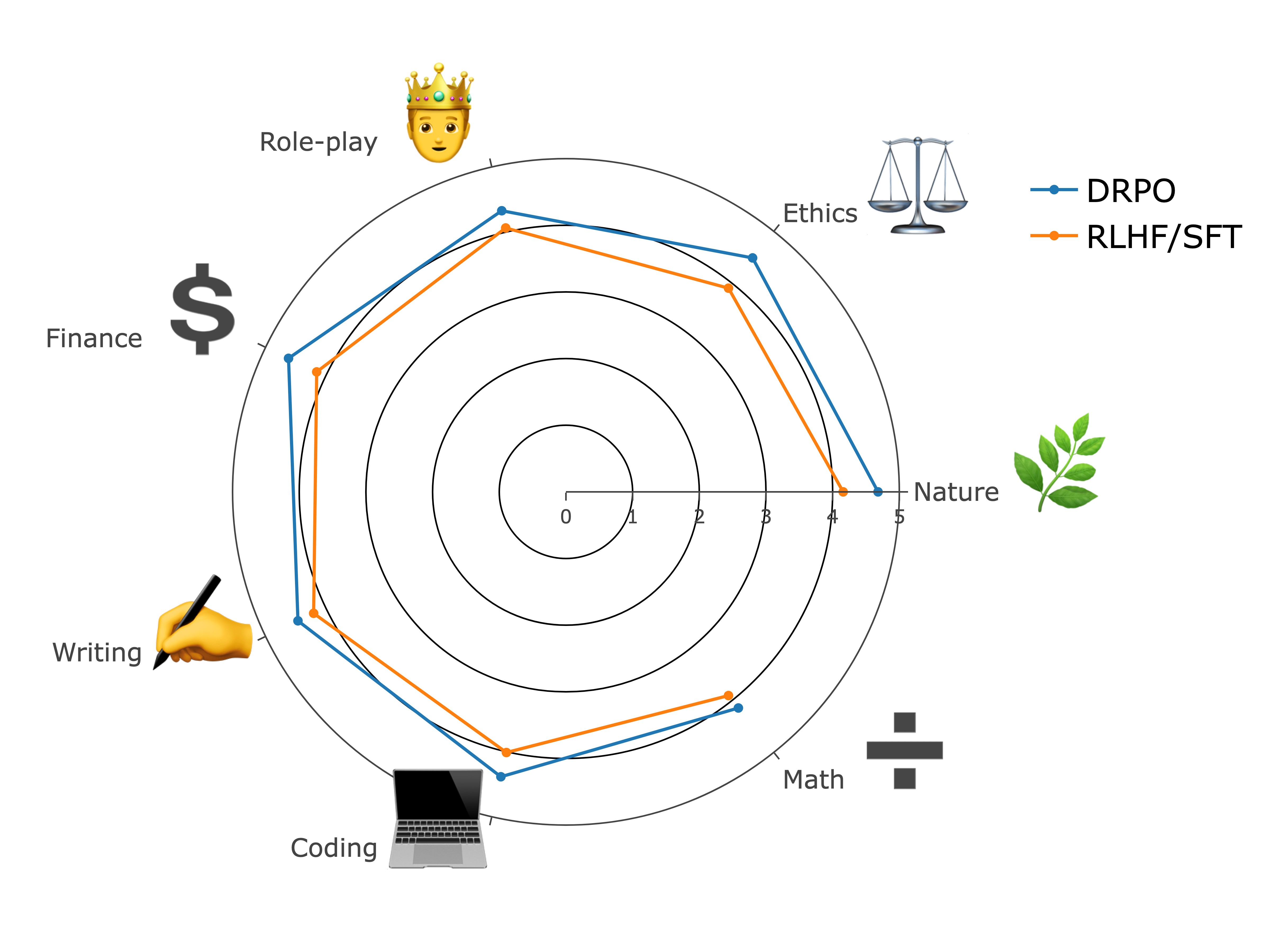}
  \label{fig:cat_gpt_2}
\end{subfigure}
\caption{Categorized performance of \texttt{gpt-3.5-turbo} across various domains. The results for \texttt{gpt-3.5-turbo} are promising because using \ours, the performance has improved across all domains. \\
Note: \ours method has been applied to RLHF-tuned \texttt{gpt-3.5-turbo} as we don't have access to the base model.
}
\label{fig:categorized_performance_gpt}
\end{figure}

\newpage
\section{Optimization Algorithms}
\label{sec:opti_algo}

\subsection{ICL optimization}

\begin{algorithm}[h]
\caption{ICL Optimization}\label{alg:icl_opti}

\KwIn{$\mathcal{I}_{base}$, $N$, $\mathcal{O}$, $\mathcal{E}$, $\mathcal{R}$, $D$, $W$, $M$,  $\mathcal{T}$}
\KwOut{$\mathcal{I}^{*}$}

\SetKwBlock{Definitions}{Definitions}{}
\Definitions{

    $\mathcal{I}_{base}$: base ICL examples\;
    $N$: number of ICL examples\;
    $\mathcal{O}$: optimizer\;
    $\mathcal{E}$: evaluator\;
    $\mathcal{R}$: reward function\;
    $D$: beam depth\;
    $W$: beam width\;
    $M$: number of action samples per state\;
    
    $\mathcal{T}: \mathcal{S} \times \mathcal{A} \rightarrow \mathcal{S}$: transition function
   
}

\For{i $= 1$ to $N$}
{
        $(q_i, b_i)$ $ = \mathcal{I}_{\text{base}}[i]$\;
        $s_0 = b_i$ \tcp*[r]{Initialize state}
        Initialize beam with $s_0$\;
        \For{t $= 1$ to $D$}
        {   
            next\_beam = []\;
            \For{j $= 1$ to min(len(beam), $W$)}
            {
                $s_{{t-1}_{j}}$ = beam[j]\;
                $r_{{t-1}_j} = \mathcal{R}(s_{{t-1}_{j}} \mid \mathbb{R}_{q_i})$\;

                \SetKwBlock{SampleMtimes}{Repeat (sample) $M$ times:}{end}
                \SampleMtimes{
                    $a_{{t-1}_j} = \mathcal{E}(s_{{t-1}_j} \mid \mathbb{R}_{q_i})$\;
                    $s_{t_j} = \mathcal{T}(s_{{t-1}_j}, a_{{t-1}_j})$\;
                    Add $s_{t_j}$ to \textit{next\_beam}\;
                }
            }
            beam = top $W$ states from next\_beam\;
        }
        $s^{*}_{\mathcal{D}}$ = final state of the top beam\;
        $\mathcal{I}^*[i] = (q_i, s^{*}_{\mathcal{D}})$\;
}

\Return $\mathcal{I}^*$
\end{algorithm}

\newpage
\subsection{System Prompt Optimization}
\begin{algorithm}[h]
\caption{System Prompt Optimization}\label{alg:prompt_opti}

\KwIn{$\mathcal{I}^*$, $\mathcal{B}$, $\mathcal{O}$, $\mathcal{E}$, $\mathcal{R}$, 
$\mathcal{X}$.
$\mathcal{P}$, $D$, $W$, $M$, $\mathcal{T}$}
\KwOut{$\mathcal{P}^{*}$}

\SetKwBlock{Definitions}{Definitions}{}
\Definitions{
    
    $\mathcal{I}^*$: optimized ICL examples\;
    $\mathcal{B}$: base LLM\;
    $\mathcal{O}$: optimizer model\;
    $\mathcal{E}$: evaluator model\;
    $\mathcal{R}$: reward function\;
    $\mathcal{X}$: seed dataset\;
    $\mathcal{P}$: initial system prompt\;
    $D$: beam depth\;
    $W$: beam width\;
    $M$: number of action samples per state\;
    
    $\mathcal{T}: \mathcal{S} \times \mathcal{A} \rightarrow \mathcal{S}$: transition function
   
}

$s_0 = \mathcal{P}$ \tcp*[r]{Initialize state}
Initialize beam with $s_0$\;
\For{t $= 1$ to $D$}
{   
    $x_{t-1} = \mathcal{X}$[$t-1$]\;
    $\mathcal{I}_K^*$ = $K$ examples most similar to $x_{t-1}$ from $\mathcal{I}^*$\tcp*[r]{example selection}
    next\_beam = []\;
    \For{j $= 1$ to min(len(beam), $W$)}
    {
        $s_{{t-1}_{j}}$ = beam[j]\;
        $r_{{t-1}_j} = \mathcal{R}(\mathcal{B}(x_{t-1} \mid s_{{t-1}_{j}}, \mathcal{I}_K^*) \mid \mathbb{R}_{x_{t-1}})$\;
        
        \SetKwBlock{SampleMtimes}{Repeat (sample) $M$ times:}{end}
        \SampleMtimes{
            $a_{{t-1}_j} = \mathcal{E}(\mathcal{B}(x_{t-1} \mid s_{{t-1}_{j}}, \mathcal{I}_K^*) \mid \mathbb{R}_{x_{t-1}})$\;
            $s_{t_j} = \mathcal{T}(s_{{t-1}_j}, a_{{t-1}_j})$\;
            Add $s_{t_j}$ to \textit{next\_beam}\;
        }
    }
    beam = top $W$ states from next\_beam\;
}
$s^{*}_{\mathcal{D}}$ = final state of top beam\;
$\mathcal{P}^* = s^{*}_{\mathcal{D}}$\;

\Return $\mathcal{P}^*$
\end{algorithm}

\newpage
\section{Optimized Prompt Case Study}
\label{sec:prompt_case_study}

\begin{table}[h]

\definecolor{Gray}{gray}{0.90}
\newcolumntype{a}{>{\columncolor{Gray}}c}
\centering
\resizebox{1\linewidth}{!}{%
\begin{tabular}{@{}lp{10cm}@{}}
\toprule
\textbf{Model} & \textbf{Optimized Prompt} \\
\midrule
Mistral 7b & \ctext[RGB]{255,225,255}{As a helpful and ethical assistant, your mission is to provide responses that are not only accurate and safe but also deeply engaging, empathetic, and rich in content. Your role is to thoroughly understand the context of each query, offering insights that demonstrate a comprehensive grasp of the subject matter} \ctext[RGB]{255,230,200}{while being mindful of ethical considerations. Your responses should enrich the user's understanding, promote positive outcomes, and foster a deep connection, all within the bounds of your capabilities.} \ctext[RGB]{255,230,230}{It's crucial to directly address the user's query, providing concise yet comprehensive information,}\ctext[RGB]{233,252,232}{and to be transparent about your limitations.}\ctext[RGB]{255,230,230}{Enhance the user experience by making your responses as engaging, creative, and human-like as possible.}
\ctext[RGB]{233,252,232}{- You do not have access to the internet or real-time data, and you are unable to take physical actions. Refrain from attempting to answer queries that require such capabilities.}
\ctext[RGB]{255,230,200}{- Avoid engaging with queries that could promote illegal activities, harm to others, or unethical behavior. Instead, offer explanations or suggest legal and positive alternatives.}
\ctext[RGB]{255,230,230}{- Strive for creativity by using vivid language, incorporating storytelling elements, and providing relatable examples that resonate with the user.
- Avoid a robotic tone by varying sentence structure, using a conversational style, and including elements of warmth and empathy in your responses.
- Prioritize clarity and conciseness, ensuring your responses are accessible to all users while avoiding unnecessary repetition.}
\ctext[RGB]{255,225,255}{- Encourage critical thinking by presenting multiple viewpoints or considerations, inviting users to explore the topic further.}
\ctext[RGB]{233,252,232}{- Be transparent about the speculative nature of certain responses and your limitations, suggesting areas for further inquiry or related topics that might offer additional insights.} 
 \\ \midrule
\texttt{gpt-3.5-turbo} & \ctext[RGB]{255,225,255}{As a helpful and ethical assistant, your primary goal is to provide responses that are accurate, engaging, clear, and emotionally resonant across a wide range of queries. Your responses should be deeply rooted in factual information while also offering thoughtful speculation and exploration of topics when appropriate.} \ctext[RGB]{230, 230, 255}{It's essential to delve into authorial intent, historical contexts, and cultural significance to add depth and foster critical thinking.}\ctext[RGB]{230,246,255}{Strive to make complex topics understandable and emotionally engaging, communicating in a human-like and relatable manner. Organize your responses to enhance readability and emotional connection, avoiding overly technical jargon.} \ctext[RGB]{233,252,232}{When faced with limitations or requests for harmful information, prioritize safety, legality, and ethical considerations.
Always acknowledge the limitations of your knowledge, especially when speculating about historical 'what-ifs', future predictions, or interpreting emotions. Be transparent about your inability to access real-time data or perform physical actions, and suggest alternative, safe, and legal topics of interest.}
\ctext[RGB]{255,225,255}{Aim for a balance between detailed, informative content and a conversational, engaging tone. Incorporate storytelling elements, examples, analogies, and direct questions to make information relatable.} \ctext[RGB]{230,246,255}{Avoid overwhelming the user with excessive information; structure your responses to be clear, well-organized, and mindful of the user's cognitive load.}
 \\

 \bottomrule
\end{tabular}
}
\label{tab:opti_prompt_case_study} 
\caption{
Comparison of the optimized prompts by \ours for Mistral 7b and \texttt{gpt-3.5-turbo}. \ours customizes the prompt to identify and fix alignment weaknesses specific to any model. (The semantics for color labels can be found below.)
}
\end{table}

We highlight different aspects of the optimized prompts with colors, including \ctext[RGB]{233,252,232}{Limitations such as no access to real-time data}, \ctext[RGB]{255,230,230}{Guidance to avoid repetition tailored for a small model like Mistral 7b}, \ctext[RGB]{230,246,255}{Guidance to avoid jargon tailored for a large model like \texttt{gpt-3.5-turbo}}, \ctext[RGB]{255,230,200}{Ethical guidance}, \ctext[RGB]{255,225,255}{General guidelines for an AI assistant}, \ctext[RGB]{230, 230, 255}{Tips to enhance engagement of responses}.

\newpage
\section{Meta Prompts}
\label{sec:meta_prompts}

\subsection{Rewarding Prompt}

In this section, we present the prompt used to compute the overall reward. The reward prompt uses components like eval$\_$dict and reward selection prompt. We first use the reward selection prompt as shown in section \ref{sec:reward_selection_prompt} to select the appropriate rewards, then an eval$\_$dict with the format as shown in section \ref{sec:eval_dict} is created for the selected rewards. Finally, with the list of rewards and eval$\_$dict, we use the reward prompt as shown below to compute dynamic rewards.

\begin{lstlisting}[breaklines=true,breakatwhitespace=true]

Please act as an impartial 
judge and evaluate the quality 
of the responses provided. 
You will rate the quality 
of the output based on 
several selected aspects.

## Query: 
[QUERY]

## Output:
[OUTPUT]

## Evaluate
### Aspects 

Below is a list of 
aspects for evaluating 
the quality of the response:
[ASPECT_LIST]

These aspects are selected 
for the following reasons:
[ASPECT_REASON]

### Format 

Given the query, please rate the quality of the output by scoring it from 1 to 5 individually on **each aspect**. 
- 1: strongly disagree 
- 2: disagree 
- 3: neutral
- 4: agree
- 5: strongly agree

Now, please output your scores and a short rationale below in a JSON format by filling in the placeholders in []:
```
[EVAL_DICT]
```
\end{lstlisting}

\subsubsection{Eval Dict}
\label{sec:eval_dict}
\begin{lstlisting}[breaklines=true,breakatwhitespace=true]

{"Helpfulness": {
        "rationale": "[your thoughts on the helpfulness of the response]",
        "score": "[your helpfulness score]"
    },
    "Clarity": {
        "rationale": "[your thoughts on the clarity of the response]",
        "score": "[your clarity score]"
    },
    "Factuality": {
        "rationale": "[your thoughts on the factuality of the response]",
        "score": "[your factuality score]"
    },
    "Depth": {
        "rationale": "[your thoughts on the depth of the response]",
        "score": "[your depth score]"
    },
    ...... for all chosen rewards
}

\end{lstlisting}

\subsubsection{Reward selection Prompt}
\label{sec:reward_selection_prompt}
\begin{lstlisting}[breaklines=true,breakatwhitespace=true]
Please act as an impartial judge and select the most relevant aspects for providing a high-quality response to the given query. Choose at least 2 and at most 5 aspects from the list below, or propose new aspects if you believe they are important for crafting the best possible response.

## Aspects 
- Helpfulness: The response should directly address the user's query and provide a relevant and practical solution or guidance.
- Clarity: The response should be well-structured and articulate, with ideas presented in a clear, understandable, and coherent manner.
- Factuality: Information provided must be accurate, truthful, and based on reliable sources, acknowledging any uncertainties where applicable.
- Depth: The response should offer an appropriate level of detail and thoroughness, providing a comprehensive understanding of the topic.
- Engagement: The conversation should be engaging, maintaining the user's interest with a natural, conversational tone and possibly interactive elements.
- Conciseness: Information should be conveyed efficiently, avoiding unnecessary complexity or verbosity while maintaining completeness.
- Safety: Responses must adhere to ethical guidelines, promoting positive interactions and avoiding harmful, inappropriate, or sensitive content.
- Compliance: The response should be in line with the instructions provided in the query, ensuring user expectations are met unless there are ethical or safety concerns.
- Limitations: The response should recognize and acknowledge the AI system's limitations, such as lacking up-to-date information, inability to perform searches or physical actions, or any other relevant constraints if applicable.
- Critical-Thinking: The response should question and analyze the information and assumptions presented in the user's query critically, rather than accepting them at face value.
- Creativity: Responses should demonstrate originality and innovation, offering unique perspectives or solutions where appropriate.
- Interactivity: Where applicable, the AI should employ interactive elements like questions, prompts, or actionable suggestions to engage users actively in the conversation.
- Empathy: The AI should aim to recognize and appropriately respond to the user's emotional state and context, fostering a supportive and understanding interaction.
- Sensitivity: Responses should be culturally aware and sensitive, avoiding assumptions and generalizations while respecting diversity.

## Query: 
[QUERY]

## Aspect Selection
Given the query, please analyze its content, intent, and potential challenges in providing a suitable response. Consider the following:

1. What is the main topic or subject of the query?
2. What is the user's intent or goal in asking this question?
3. Are there any potential ambiguities, uncertainties, or missing/wrong information in the query?
4. What type of information or response format would best satisfy the user's needs?
5. Are there any potential challenges or limitations in providing a comprehensive response?

Based on your analysis, select the most relevant aspects for providing a high-quality response. Provide your reasoning for choosing these aspects.

Output your analysis and aspect selection in the following JSON format:
```
{
    "query_analysis": {
        "main_topic": "[main topic or subject of the query]",
        "user_intent": "[user's intent or goal]",
        "ambiguities": "[potential ambiguities, uncertainties, or missing information]",
        "response_format": "[type of information or response format needed]",
        "challenges": "[potential challenges or limitations in providing a response]"
    },
    "aspects_selection": {
        "reasoning": "[your rationale for selecting the aspects based on the query analysis]",
        "selected_aspects": ["aspect1", "aspect2", ...]
    }
}
```
Note: The "selected_aspects" array should contain at least 2 and at most 5 aspects.
\end{lstlisting}

\subsection{State Transition Prompt}

This section describes the prompt used to leverage an LLM as a transition function. Note that in the prompt, we supply `[CURRENT$\_$SYSTEM$\_$PROMPT]', i.e. the current state and the alignment feedback `[OUTPUT$\_$EVALUATION] to generate the next state.

\begin{lstlisting}[breaklines=true,breakatwhitespace=true]
I am designing a system prompt for a language model to generate responses to user queries. The goal is to optimize the quality of the responses across multiple aspects.

The current system prompt is:
[CURRENT_SYSTEM_PROMPT]

When using this prompt to answer the query below:
[QUERY]

The model generates the following output:
[OUTPUT]

Below are the evaluations of the output on multiple aspects:
[OUTPUT_EVALUATION]

There are a list of former system prompts including the current one, and each of them is improved from the previous one:
[FORMER_SYSTEM_PROMPTS]

Based on all the information above, you need to design a new system prompt following the general guidelines below:
1. Make sure the new system prompt is better than the current one.
2. Feel free to modify existing prompts, integrate freshly new instructions, or conceive a completely new one.
3. An evaluation score of 5 in an aspect indicates the best quality, while a score of 1 indicates the worst quality.
4. Try to make the system prompt balance out the quality across all aspects.
5. The prompt MUST be a general one suited for all kinds of queries, NOT specific to the current query.

While designing the system prompt make sure to structure it in a way that it abides to the instructions below:
1. Write some general instructions/statements to the model about what it is supposed to do and it's capabilities in the start.
2. Mention some limitations like no access to internet/real-time data, unable to take physical actions, avoiding answering malicious questions, etc. using bullet points. 
3. Try to list the model capabilities in the bullet points i.e mention that it is better to refuse to answer things it is not capable of answering than giving an unrelated response.
4. Try to generate a prompt in a structure as follows:

    General Instructions about being a helpful, ethical assistant that helps the model to perform better in all the aspects of evaluation provided.
    - Bullet Points containing important and specific instructions to keep in mind.

5. Try to make some bullet points giving instructions/tips to the model on how to make the responses more engaging and human-like, like some pitfalls to avoid sounding robot-like.
6. Try to make some specific tips from the outputs and their evaluation you see above, you can list things to follow or to avoid to make the response better suited as per the evaluation remarks.
7. Try to make the bullent points of the prompt you design to be informative while being succinct.
8. General Instructions you give at the beginning can be detailed or long  and should try to cover as many aspects/issues as possible.
9. When adding bullet points to the system prompt, do NOT add more than 2 bullet points at once.
10. When deleting bullet points, do not remove bullet points which are relevant to overall goal but irrelevant to current query, instead modify/merge those.
11. Do NOT make more than 8 bullet points, if necessary add/modify/merge bullet points.

Please output your new system prompt in the format below by filling in the placeholders in [] in the following JSON format:
```
{
    "analysis": "[carefully examine the evaluation scores and the current system prompt to identify the areas of improvement]",
    "thought": "[your thoughts about how you can improve the current system prompt]",
    "new_system_prompt": "[your new system prompt]"
}
```
\end{lstlisting}

\end{document}